\documentclass[times,twocolumn,final]{elsarticle}

\usepackage{medima}
\usepackage{framed,multirow}
\usepackage{afterpage}

\usepackage{amssymb}
\usepackage{latexsym}

\usepackage{url}
\usepackage{xcolor}
\usepackage{booktabs,caption}
\usepackage{threeparttable}
\usepackage{amsmath}
\usepackage{float}
\usepackage{graphicx}
\usepackage{subfig}
\usepackage{tabularx}

\usepackage{hyperref}

\newcommand{\revision}[1]{{\color{black}#1}}
\newcommand{\ie}[0]{{i.e.}}
\newcommand{\eg}[0]{{e.g.}}

\newcommand{\ours}{{\texttt{FedDetect}}}

\definecolor{newcolor}{rgb}{.8,.349,.1}

\journal{Medical Image Analysis}

\begin{document}

\verso{Jin R. and Li, X.}

\begin{frontmatter}

\title{Backdoor Attack and Defense in Federated Generative Adversarial Network-based Medical Image Synthesis}%

\author[1]{Ruinan \snm{Jin}}
\author[2]{Xiaoxiao \snm{Li}\fnref{fn1}\corref{cor1}}
\cortext[cor1]{Corresponding author: Xiaoxiao Li \ead{xiaoxiao.li@ece.ubc.ca}}
\address[1]{Computer Science Department, The University of British Columbia, BC, V6T 1Z4, Canada}
\address[2]{Electrical and Computer Engineering Department, The University of British Columbia, BC, V6T 1Z4, Canada}


\begin{abstract}
Deep Learning-based image synthesis techniques have been applied in healthcare research for generating medical images to support open research and augment medical datasets. Training generative adversarial neural networks (GANs) usually require large amounts of training data. Federated learning (FL) provides a way of training a central model using distributed data while keeping raw data locally. However, given that the FL server cannot access the raw data, it is vulnerable to backdoor attacks, an adversarial by poisoning training data. Most backdoor attack strategies focus on classification models and centralized domains. \emph{It is still an open question if the existing backdoor attacks can affect GAN training and, if so, how to defend against the attack in the FL setting.} In this work, we investigate the overlooked issue of backdoor attacks in federated GANs (FedGANs). The success of this attack is subsequently determined to be the result of some local discriminators overfitting the poisoned data and corrupting the local GAN equilibrium, which then further contaminates other clients when averaging the generator's parameters and yields high generator loss. Therefore, we proposed \ours{}, an efficient and effective way of defending against the backdoor attack in the FL setting, which allows the server to detect the client's adversarial behavior based on their losses and block the malicious clients. Our extensive experiments on two medical datasets with different modalities demonstrate the backdoor attack on FedGANs can result in synthetic images with low fidelity. After detecting and suppressing the detected malicious clients using the proposed defense strategy, we show that FedGANs can synthesize high-quality medical datasets (with labels) for data augmentation to improve classification models' performance. 
\end{abstract}


\end{frontmatter}



\section{Introduction}


\label{sec1}While deep learning has significantly impacted healthcare research, its impact has been undeniably slower and more limited in healthcare than in other application domains. A significant reason for this is the scarcity of patient data available to the broader machine learning research community, largely owing to patient privacy concerns. Although healthcare providers, governments, and private industry are increasingly collecting large amounts and various types of patient data electronically that may be extremely valuable to scientists, they are generally unavailable to the broader research community due to patient privacy concerns. Furthermore, even if a researcher is able to obtain such data, ensuring proper data usage and protection is a lengthy process governed by stringent legal requirements. This can significantly slow the pace of research and, as a result, the benefits of that research for patient care. 

Synthetic datasets of high quality and realism can be used to accelerate methodological advancements in medicine~\citep{dube2013approach,buczak2010data}. While there are methods for generating medical data for electronic health records~\citep{dube2013approach,buczak2010data}, the study of medical image synthesis is more difficult as medical images are high-dimensional.
With the growth of generative adversarial networks (GAN)~\citep{goodfellow2014generative}, high-dimensional image generation became possible. Particularly, conditional GAN~\citep{mirza2014conditional} can generate images conditional on a given mode, e.g. constructing images based on their labels to synthesize labeled datasets.  In the field of medical images synthesis, ~\cite{teramoto2020deep} proposed a progressive growing conditional GAN to generate lung cancer images and concluded that synthetic images can assist deep convolution neural network training. ~\cite{yu2021generative} recently used conditional GAN-generated pictures of aberrant cells in cervical cell classification to address the class imbalance issue. 
In addition, a comprehensive survey about the role of GAN-based argumentation in medical images can be found in~\cite{shorten2019survey}.

Like most deep learning (DL)-based tasks, limited data resources is always a challenge for GAN-based medical synthesis. In addition, large, diverse, and representative dataset is required to develop and refine best practices in evidence-based medicine. However, there is no single approach for generating synthetic data that is adaptive for all populations~\citep{chen2021synthetic}. Data collaboration between different medical institutions (of the heterogeneity of phenotypes in the gender, ethnicity, and geography of the individuals or patients, as well as in the healthcare systems, workflows, and equipment used) makes effects to build a robust model that can learn from diverse populations. But data sharing for data collection will cause data privacy problems which could be a risk of exposing patient information~\citep{malin2013biomedical, scherer2020joint}. Federated learning (FL)~\citep{konevcny2016federated} is a privacy-preserving tool, which keeps data on each medical institute (clients) locally and exchanges model weights with the server to learn a global model collaboratively. As no data sharing is required, it is a popular research option in healthcare~\citep{rieke2020future, usynin2022can}. Federated GAN (FedGAN) is then proposed to train GAN distributively for data synthesis~\cite{augenstein2019generative, rasouli2020fedgan}. \emph{Its overall robustness against attacks is under explored}.

\begin{figure}[H]
    \centering
    \includegraphics[width=0.4\textwidth]{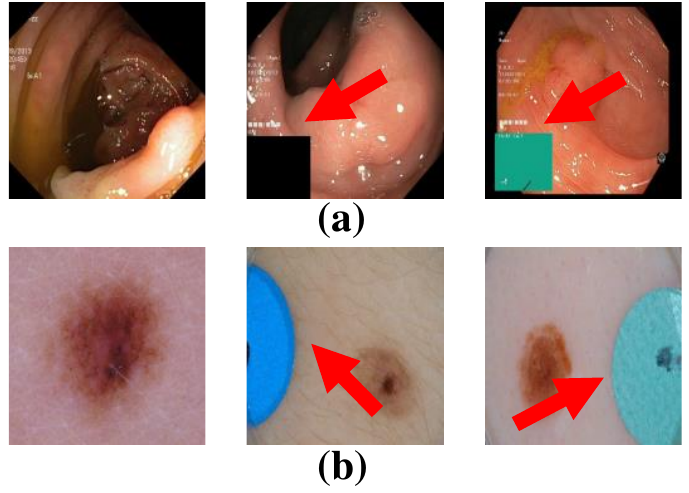}
    \caption{Example medical images with backdoor-alike noisy patches from (a)KVASIR dataset~\citep{pogorelov2017kvasir} (b) ISIC dataset~\citep{codella2018skin}}
    \label{fig:medical_noisy}
\end{figure}

However, as an open system, FL is vulnerable to malicious participants and there are already studies deep dive into different kinds of attacks for classification models in federated scenarios, like gradient inversion attacks~\citep{huang2021evaluating}, poisoning and backdoor attacks~\citep{bagdasaryan2020backdoor}. In a backdoor attack with classification models, the attacker, such as a malicious client, adds a "trigger" signal to its training data and identifies any image with a "trigger" as other classes~\citep{saha2020hidden}.
A "trigger," such as a small patch with random noise, could lead a sample to be misclassified to another class. This kind of attack takes advantage of the classification model's tendency to overfit the trigger rather than the actual image~\citep{bagdasaryan2020backdoor}.
It is worth noting that numerous medical images naturally exhibit backdoor-like triggers and noisy labels, as depicted in Fig.~\ref{fig:medical_noisy}, which enhances the significance of backdoor studies in the field of medical image analysis~\citep{xue2022robust}. Unfortunately, the server cannot directly detect the attacked images given the decentralized nature of FL, where the clients keep their private data locally. Recently, several studies found backdoor attached clients can cause a substantial drop in the classification performance in FL~\citep{tolpegin2020data,sun2021data}. These facts inspire us to think about how backdoor attack affects generative models in FL.

In this work, we focus on backdoor attacks in the labeled medical image synthesis using conditional FedGAN. In our experiments, we employ two widely used medical datasets. First, we demonstrate the effect of adding different sizes and types of the trigger (e.g., a patch with different patter and sizes ranging from 0.5 to 6.25 percent of the original image size) can significantly affect the fidelity of the generated images. Second, we propose an effective defense mechanism, \ours{}, to detect malicious client(s). We observe that the attacked discriminator tends to overfit, yielding inconsistent and different training loss patterns. Therefore, 
\ours{} performs outlier detection on clients' shared training loss in each iteration and red flags a client as malicious if it is detected as an outlier based on its record in multiple iterations. Our results first qualitatively compare the visualizations of the generative images of \ours{} and with the alternative defense strategies under backdoor attacks with different trigger sizes.  Furthermore, with the FedGAN-assisted diagnostic as a signifier for FedGAN's role in the field of medicine, we quantitatively evaluate the effect of FedGAN-assisted data augmentation for DL training, the attack, and \ours{}. It shows that FedGAN-assisted data augmentation effectively improves the performance of DL training, while synthetic medical images generated from the attacked FedGANs corrupt their utility, leading to a poor medical utility value. \ours{} blocks such adversarial and builds up a Byzantine-Robust FedGAN~\citep{fang2020local}.

Our contributions are summarized as follows:
\begin{itemize}
    \item To the best of our knowledge, we are the first to examine the robustness of FedGAN, a promising pipeline for medical data synthesis, from the practical backdoor attack perspectives. Without loss of generality, we examine our proposed pipeline, attack, and defense on two public medical datasets. 
    \item We propose a general pipeline of conditional FedGAN to generate labeled medical datasets. We extend backdoor attacks for classification models to generative models and reveal the vulnerability of conditional FedGAN training by backdoor attacks. We investigate the effect of different trigger sizes and types in the attacks.  
    \item
    We propose an effective defense strategy \ours{} and compare it with the current practices to show its irreplaceable role in achieving robust FedGANs. We not only present qualitative results by examining the fidelity of the synthetic data, but also quantitatively evaluate their utility as data augmentation to assist diagnostic model training.
    We show the practical use and the innovation of \ours{} in the field of medical image synthesis.
\end{itemize}

A preliminary version of this work, Backdoor Attack is a Devil in Federated GAN-based Medical Image Synthesis~\citep{jin2022backdoor} has been presented in SASHIMI 2022. This paper extends the preliminary version by expanding the conditional FedGAN which generates synthetic medical images with labels so that synthetic images can serve for broader usage (such as data augmentation), examining the effect of various sizes and types of triggers with gray-scale and RGB medical datasets, performing quantitative analysis on the synthetic data, and assessing the utility of the synthetic medical images as data augmentation to assist training deep diagnostic models.

\section{Preliminaries and Related Work}
\subsection{Conditional Generative Adversarial Networks}
\label{rw:GAN}
GAN was first been proposed in~\cite{goodfellow2014generative} as an unsupervised generative algorithm, where two deep neural networks, discriminator and generator are training against each other to optimize minmax objective function Eq.~\ref{eq:ganloss}.
\begin{equation}
\label{eq:ganloss}
\min_{G}\max_{D}\mathbb{E}_{x\sim p_{\text{data}}(x)}[\log{D(x)}] +  \mathbb{E}_{z\sim p_{\text{z}}(z)}[\log{(1 - D(G(z)))}],
\end{equation}
where $G$ and $D$ are generator and discrimnator, $x$ is the training data, and $z$ is a random noise vector sampled from a predefined distribution $p_z.$ GAN has been used to generate medical image datasets for data augmentation and data sharing for open research as healthcare institutions are regulated to release their collected private data~\citep{chen2022generative, lin2022insmix}.

Later,~\cite{mirza2014conditional} implemented the conditional GAN, turning GAN to be supervised learning algorithm, where both the discriminator and the generator take an extra auxiliary label so that GAN generates images conditional on the given label according to updated objective function Eq.~\ref{eq:cganloss1}. 
\begin{equation}
\label{eq:cganloss1}
\min_{G}\max_{D}\mathbb{E}_{x\sim p_{\text{data}}(x)}[\log{D(x|c)}] +  \mathbb{E}_{z\sim p_{\text{z}}(z)}[ \log{(1-D(G(z|c)))}],
\end{equation}
where we add class label $c$ as the conditional term compared to Eq.~\ref{eq:ganloss}. Generating synthetic medical data using conditional GAN is gain more practical values in healthcare scenarios, because medical data is usually labeled and this label makes it meaningful for diagnostic purposes, e.g., for classifying if certain patient has the disease~\citep{frangi2018simulation}, and for data augmentation~\citep{chen2022generative}.


\subsection{Federated Learning}
\label{rw:fl}
Training DL models usually requires a large amount of training data. However, collecting data is challenging in the field of healthcare because healthcare providers, governments, and related medical organizations must pay particular attention to the patient's privacy and guarantee the proper use of their collected data~\citep{price2019privacy}. In this case, limited data in local healthcare institutions is usually
biased and unilateral~\citep{wang2020federated}, which in turn impede the AI-assisted diagnostic technology in healthcare~\citep{van2014systematic}.

FL has been proposed as a promising strategy to facilitate collaboration among several institutions (\eg, medical centers distributed from different geographical locations) to train a global DL model~\citep{konevcny2016federated}. Given the important role of FL plays in leveraging medical data from distributed locations and the practical usage of DL-based synthetic models in medicine, combining them together will help facilitate advancement in medicine. Previous studies try to establish the FedGAN~\citep{rasouli2020fedgan} and explored its robustness in terms of the differential privacy~\citep{augenstein2019generative}. 

In addition, Byzantine-Robust FL is a key challenge in FL deployment, as the clients are barely controllable and typically viewed as an open system. Literature has shown that FL is vulnerable to multiple kinds of adversaries~\citep{bouacida2021vulnerabilities, liu2022intervention}. Example vulnerabilities includes model poisoning attacks~\citep{bhagoji2019analyzing}, gradient inversion attacks~\citep{huang2021evaluating}, inference attacks~\citep{ying2020privacy}, backdoor attacks~\citep{li2022backdoor}, etc.

\subsection{Backdoor Attack}
In this section, we begin by introducing the general concept of a backdoor attack. Next, we delve into the specific details of backdoor attacks in FL and backdoor attacks in generative models.

\paragraph{General concept}The backdoor attackers aim to embed a backdoor, also known as trigger, in the training data in order to corrupt the performance of the Deep Neural Network. Given it involves poisoning data, it belongs to the fields of data poisoning attacks, which has been widely explored in multiple machine learning fields, including Support Vector Machines, Statistical Machine Learning, and DL~\citep{biggio2012poisoning, nelson2008exploiting}. In DL, current studies mainly explore poisoning attacks in centralized classification models, where a hidden trigger is pasted on some of the training data with wrong labels. The attacker activates it during the testing time so that the classification model produces a lower testing accuracy for images with triggers~\citep{saha2020hidden}. This attack strategy takes advantage of the tendency that the deep neural network is more likely to learn the pattern of the backdoor instead of the actual image~\citep{li2022backdoor}.

\paragraph{Backdoor attack in FL}
Due to the distributed nature of FL, the server has little control on the client side. Namely, FL is even more vulnerable to backdoor and poisoning attacks. Existing studies have explored such attacks in classification models from various perspectives.~\cite{bagdasaryan2020backdoor} proposed to use apply \textit{model replacement} as a means to introduce backdoored functionality into the global model within FL.~\cite{fang2020local} introduced the initial concept of local model poisoning attacks targeting the Byzantine robustness of FL.~\cite{wang2020attack} introduced a novel concept of an \textit{edge-case backdoor}, which manipulates a model to misclassify seemingly straightforward inputs that are highly unlikely to be part of the training or test data.~\cite{sun2019can} conducts a thorough investigation of backdoor attacks and defense strategies in FL classification models

\paragraph{Backdoor attack on GAN}
The existing backdoor attack has been focusing on classification models. Despite this, the applicability of backdoor attacks against GANs is underexplored, especially in the field of medicine. This is due to the fact that backdoor attacks against GANs are more complex, since the input for GANs is a noise vector, while the output is a generated-new-mage. Backdoor in FL for classification models was initially introduced by ~\cite{bagdasaryan2020backdoor}, where a constrain-and-scale technique is applied to amplify the malicious gradient of clients. In the study, the local clients are flexible to employ local training procedures. However, this scenario is unlikely to happen in the medical circumstances, given the organizer of the FL can enforce the proper training procedure on reliable hardware to prevent malicious clients from taking adversarial actions~\citep{pillutla2019robust}. In contrast to the previous backdoor federated classification study, we assume that there is a reliable FL pipeline that each client follows the provided training rules to calculate and update the correct parameters without modifying them, thereby performing the given training process in accordance with the instructions given by the trusted server (the organizer of the FL in reality). Then we investigate how backdoor attack can affect GAN for data synthesis in this FL setting. 

\subsection{Defending Backdoor Attack}
\label{rw:defense}
There are a variety of defense strategies for backdoor attacks ranging from data level to model level but with a focus on the classification models in centralized training. Data level defenses mainly include two perspectives: 1) Detect the trigger pattern and either eliminate it from the data or add a trigger blocker to decrease the contribution of the backdoor during training the DL model~\citep{doan2020februus}. 2) Perform a series of data argumentation (e.g., shrinking, flipping) before feeding the data into the model~\citep{qiu2021deepsweep,villarreal2020confoc}. This is because the transitional backdoor attack is sensitive to the pattern of the trigger and the location of the backdoor patch~\citep{li2021backdoor}.

There are more model-level defense strategies: 1) Reconstructing attacked model strategy. This strategy aims to retrain the trained attacker model with some benign samples to alleviate the backdoor attacks. 2) Synthesis trigger defencing strategy. It attempts to perform outlier detection on the DL models by reconstructing either the specific trigger~\citep{wang2019neural, harikumar2020scalable} or the trigger distribution~\citep{zhu2020gangsweep, guo2020towards}. 3) Diagnosing attacked model through meta-classifier. This method applies certain pre-trained classifiers to identify the potentially infected models and prevent deploying them~\citep{kolouri2020universal, zheng2021topological}. 4) Unlearning the infected model. The unlearning strategy first detects the malicious behavior and then defends against the backdoor attack by performing reverse learning through utilizing the implicit hypergradient~\citep{zeng2021adversarial} or modifying the loss function~\citep{li2021anti}. 5) Robust aggregation. This strategy is specifically tailored for FL. It involves meticulous adjustments of the server aggregation learning rate, taking into account client updates on a per-dimension and per-round basis~\citep{pillutla2019robust}. 6) Adversarial distillation. This defense strategy is also designed for classification tasks in FL. It employs a GAN on the server side to acquire a distillation dataset. Subsequently, knowledge distillation is applied, utilizing a clean model as the teacher to educate the server model and remove the backdoored neurons from the malicious model~\citep{zhu2023adfl}. 7) Trigger inverse engineering. This is a provable technique that is proposed specifically for backdoor attacks under FL classification models~\citep{zhang2022flip}. 

It requires the benign clients to apply trigger inversion techniques in the training time to construct an augmented dataset that consists of poisoned images with clean labels. This serves as model hardening and reduces the prediction confidence of backdoored sample. The inference step then takes advantage of this prediction confidence to perform classification tasks. 

Moreover, the comprehensive backdoor attack surveys can be found in~\cite{li2022backdoor} and ~\cite{guo2022overview}.

Unfortunately, most of the above defense strategies do not fit the FedGAN settings, either due to the server cannot access the local private data, or the training of GAN-based generative models does not behave the same as the classification models.
\begin{figure*}[t]
	\centering
	\includegraphics[width=1\linewidth]{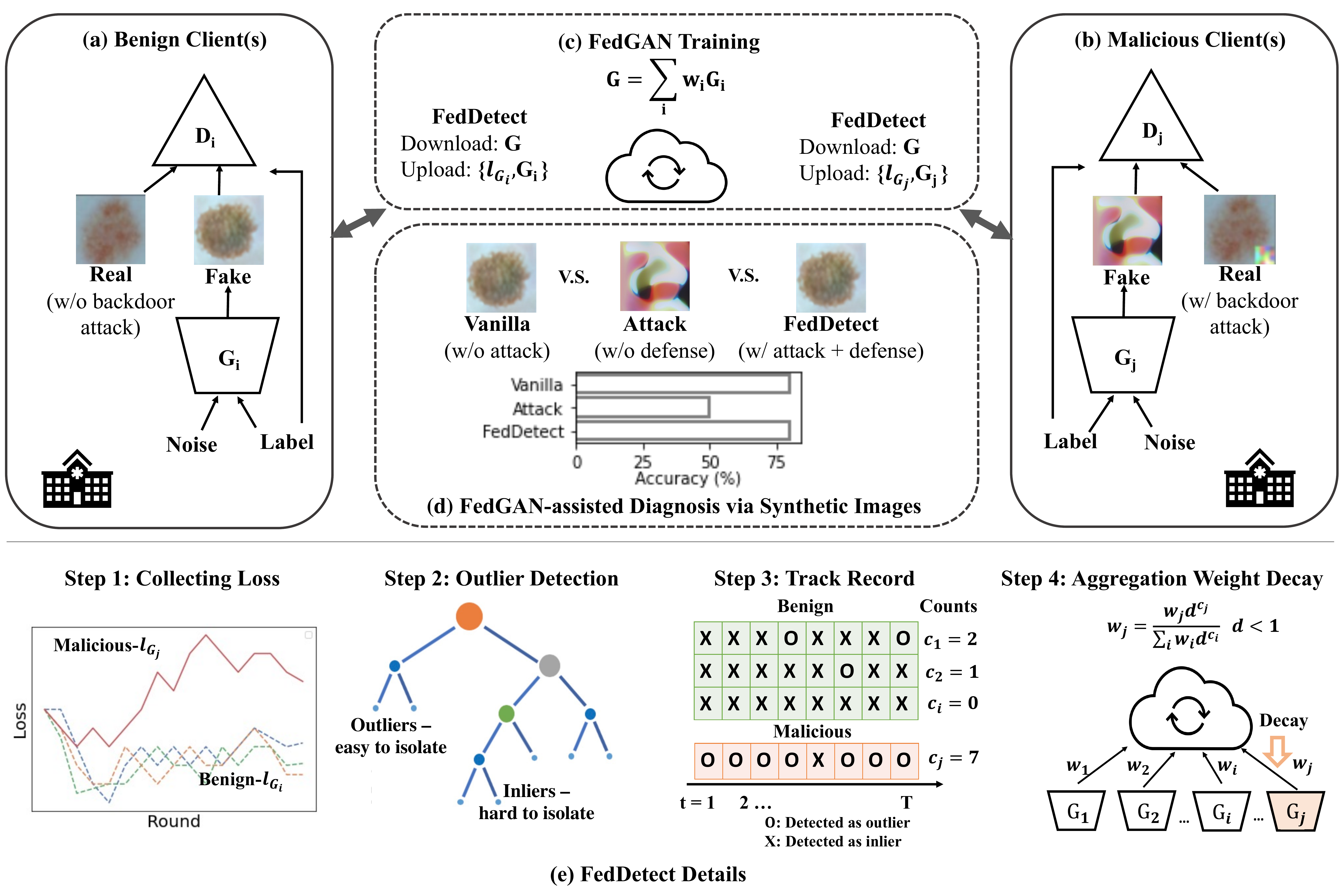}
	\caption{The overview of our proposed framework. Specifically, our FedGAN scenario consists of benign clients shown in (a), malicious clients shown in (b), and a global conditional FedGAN server as illustrated in (c). During the training, the malicious discriminator is fed with poisoned images. The discriminator and generator are trained against each other locally following the training protocol enforced by the FedGAN organizer. The generators upload their training loss and aggregate parameters per global iteration as described in (c). In the end, we expect the server generator to produce high-quality images for medical research in terms of fidelity and diversity to assist medical diagnosis tasks as demonstrated as Vanilla in (d). However, the attack degrades the central generator's performance, generating noisy images with little value for medical diagnosis. We propose \ours{} to defend against such attack. As illustrated in (e), \ours{} consists of four steps. First, the FL server collects the local generator loss and then passes them to the isolation forest for anomaly detection. Because outliers are located distantly from inliers, they are supposed to be isolated at the early stage in the construction of the isolation trees. The outliers are flagged as potential adversarial and the server tracks the record of the number of each client being flagged. Finally, the FedGAN server decays the weight of those potential malicious clients based on the track record per round. \ }
	\label{fig:framework}
\end{figure*}

\section{Methods}

The goal of medical image synthesis is to generate high-quality synthetic data that can be employed for open research to argue the limited datasets and balance the training data, and finally hasten DL methodological advancements in medicine. In our study, we explore conditional GAN in the FL setting, \ie, conditional FedGAN.

In this section, we first introduce our setting for FedGAN in Section~\ref{method:FedGAN} Next, we discuss the scope for adversarial attacks and determine the best way to implement the backdoor attack that involves data poisoning in  Section~\ref{method:attack}. Then, we suggest the potential strategies for defending against such attack in FedGAN to build a robust FL system in Section~\ref{method:defence}.

\begin{algorithm*}[t]
    \caption{\ours{}}
    \label{alg:cap}
    \textbf{Notations: }{Clients $C$ indexed by $i$; local discriminator $D_i$, and generator $G_i$, local generator loss $l_{G_i}$, global generator $G_{\rm server}$, aggregation weight $w_i \in [0,1]$; times of being detected as malicious $i$-th client $c_i$, local updating iteration $K$; global communication round $T$, total number of clients $N$, decay rate $d$, warmup iteration $m$.}
	\begin{algorithmic}[1]
	\State $c_i \leftarrow 0$, $w_i \leftarrow \frac{1}{N}$ 
	 \Comment{Initialization}
	\State{For $t = 0 \to T$, we iteratively run \textbf{Procedure A} then \textbf{Procedure B}}
	\Procedure{\textbf{A}. ClientUpdate}{$t,i$} 
	            \State $G_{i}(t, 0)\leftarrow G_{\rm server}(t)$ \Comment{Receive global generator weights update}
	            \For{$k=0 \to K-1$ } 
	                \State $D_{i}{(t,k+1)} \leftarrow \text{Optimize }\ell_D( D_{i}{(t,k)}, G_{i}{(t,k)})$
	                \Comment{Update $D$ using Eq.~\eqref{eq:discLoss}}
	                \State $G_{i}{(t,k+1)} \leftarrow \text{Optimize }\ell_G( D_{i}{(t,k+1)}, G_{i}{(t,k)})$
	                \Comment{Update $G$ using Eq.~\eqref{eq:GenLoss}}
	                \EndFor
	   \color{black}
	    \State Send $G_i(t,K)$ and $l_{G_i}$ to \textsc{ServerExecution}
	    \color{black}
	\EndProcedure
	
    \Procedure{\textbf{B}. ServerExecution} {$t$}:
                \For{each client $C_i$ \textbf{in parallel}}
\color{black}
                \State{$G_i, l_{G_i} \gets$
                \textsc{ClientUpdate}$(t,i)$}
                \Comment{Receive local model weights and loss.}
                \If{$t > m$}
                \Comment{Start detection after warmup}
                \State{$O \gets$ \textsc{IsolationForest}$(\ell_{G_1}...\ell_{G_N})$}
                
                \If{$0 < |O| < \frac{1}{2}|N|$} \Comment{Detect valid number of outliers}
    \For{each detected client  $C_i$ \textbf{in} $O$}
    \State{$c_i \leftarrow c_i + 1$}
    \Comment{Increment total count $C_i$ been detected as outlier}
    \State{$w_i \leftarrow w_i \times d^{c_i}$}
    \Comment{Decay weights for outliers}
    \EndFor
\EndIf
\EndIf
        \EndFor
        \State $w_i \gets \frac{w_i}{\sum_{i \in [N]} {w_i}}$
        \Comment{Normalize weights}
           	        \State $ G_{\rm server}(t+1) \leftarrow \sum_{k \in [N]} w_k  G_i(t)$
            \Comment{Aggregation on server}
        \color{black}
        \EndProcedure
	\end{algorithmic}
\end{algorithm*}

\subsection{Federated Generative Adversarial Network} 
\label{method:FedGAN}
Motivated by the local model poisoning attacks in Byzantine-robust FL classification models proposed in~\cite{fang2020local}, we depict the framework of FedGAN using a commonly FL training strategy that averages the shared model parameters, FedAvg~\citep{mcmahan2017communication}, in Fig.~\ref{fig:framework}. As introduced in Section~\ref{rw:fl} that studies have extensively explored gradient inversion attacks and inference attacks to infer or reconstruct the inputs of FL models given the shared model parameters, exposing FL training in privacy risk. In light of this, we keep the discriminators (binary classification neural networks) in GAN updating locally as they have direct access to clients' private data, making them vulnerable to data leakages if their gradients are inverted or inference attacks are used in FL training~\citep{huang2021evaluating}. Our FedGAN framework only exchanges generator's parameters with the server and uses FedAvg to update the global generator. To this end, our FedGAN locally trains both discriminator and generator pairs and globally shares generators' parameters, which is motivated from~\cite{chang2020synthetic}.

Formally, we assume that a trusted central generator $G_{\rm server}$ synthesizes images from a set of $N$ federated clients.  Each client $C_i $, for $ i \in [N]$ consists a locally trained discriminator $D_i$, and a generator $G_i$. $G_i$ takes random Gaussian noise $z$  and conditional label $\tilde{y}$ as inputs to generate synthetic images, and $D_i$ distinguish the synthetic image $\tilde{x}=G(z, \tilde{y})$ v.s. private image $x$ and its real label $y$. 
We keep $D_i$ locally and adopt FedAvg~\cite{mcmahan2017communication} to aggregate $G_i$ to $G_{\rm server}$, namely,
\begin{equation}\label{eq:fedavg}
    G_{\rm server} = \sum_{i=1}^N w_iG_i,
\end{equation}
where $w_i$ is a coefficient for weighted averaging. 
At the end, our federated GAN generate synthetic medical data $G_{\rm server}(z, \tilde{y}) \sim p_{\rm data}$ on the server side.

As a valid assumption in the FL system among medical centers, we assume every client, including those malicious ones, follows the given training protocol. For instance, they compute gradients correctly as instructed by the server and update the exact parameters when needed. By requiring FL computations to be carried out on trusted hardware, this can be accomplished~\citep{pillutla2019robust}.

\subsection{Backdoor Attack Strategies}
\label{method:attack}
Backdoor attack is a training time attack that embeds a backdoor into a model by poisoning training data (\eg, adding triggers on the images). The state-of-the-art backdoor attacks focus on image classification model~\citep{chen2017targeted,liu2020reflection} and has been recently studied on FL~\citep{bagdasaryan2020backdoor}. Current studies of backdoor attacks in deep generative models (\eg, GAN) train on a poisoned dataset~\citep{rawat2021devil} or input a poisoned noise vector~\citep{salem2020baaan} into the generator so that GAN failed to produce quality synthetic images with similar distribution as the real data. As injecting backdoor to images has been shown more effective~\citep{rawat2021devil}, we suggest a way of attacking federated GAN only through the poisoned data with more details below.

\paragraph{Adversarial Goals} 
The objective of the attacker performing a backdoor attack is to corrupt the  Byzantine-robust FedGAN ~\citep{cao2020fltrust}, leading to model divergence or converging to sub-optimal. In FL settings, adversarial clients attack the server generator using poisoned images as the input of discriminator so that the generator can be fooled, no longer generating fake medical-valued images $\tilde{x}$ with high fidelity. That is, $p_{(\tilde{x}|y)} \neq p_{(x|y)}$, for $x$ as real image and $y$ as the corresponding real label.

\paragraph{Adversarial Capabilities} As mentioned in Section~\ref{method:FedGAN} that a robust FL pipeline includes a trusted server (the organizer of the FL), which has control over the local training process. Thus, the only room for attack is through providing poisoned data to the local discriminator as shown in Fig.~\ref{fig:framework}.

\paragraph{Adversarial Motivation} A vanilla GAN optimizes loss function in the manner outlined in \cite{goodfellow2014generative}, where the discriminator seeks to maximize the accuracy of the real and fake image classification while the generator seeks to minimize the likelihood that its generated image will be classified as fake. We assume the data of all the clients are from the same distribution, \ie. $x\sim p_{\rm data}(x)$. We sample noise vectors $z$ from multivariate Gaussian distribution $p_{\rm z}(z)$. Specifically, the objective is written as follows. In each iteration, for each local discriminator $D_i$ with $i \in [N]$ as the index of $i$-th client, we perform the local optimization for updating discriminators model parameters when global generator $G$'s parameters are fixed:
\begin{equation}
\label{eq:discLoss}
\max_{D_i}\mathbb{E}_{x\sim p_{\text{data}}(x)}[\log{D_i(x_i|y_i)}] +  \mathbb{E}_{z\sim p_{\text{z}}(z)}[ \log{(1-D_i(G(z_i|y_i)))}].
\end{equation}
Then, we optimize the generator $G_i$ with $D_i$ fixed by
\begin{equation}
\label{eq:GenLoss}
\sum_{i=1}^{N}w_i\min_{G_i}\mathbb{E}_{z\sim p_{\text{z}}(z)}[ \log{(1-D_i(G_i(z_i|y_i)))}],
\end{equation}
then aggregate the global generator $G$ using FedAvg~\citep{mcmahan2017communication} as shown in Eq.~\eqref{eq:fedavg}.

The optimization of GAN is recognized to be difficult, nevertheless, because the generator is subpar upon learning that $log(D(G(z|y)))$ is probably overfitting given the fact that the discriminator is easy to fit~\citep{goodfellow2014generative}. Based on the observation and motivated by the backdoor attacks that fool the classifier to memorize (\ie, overfit) the triggers in the classification settings, we implement the trigger following the overfitting principle into the discriminator of FedGAN's training. The following part gives a detailed explanation of our adversarial model.

\paragraph{Adversarial Model} Following the realistic setting discussed in~\cite{bhagoji2019analyzing}, we assume that the malicious clients are the minority of the FL participants since the purpose of FL is to augment the training data and it is meaningless to lose trust among the majority of clients. Formally, our threat model contains a set of $M$ adversarial clients, where $|M| = \alpha |N|$ and $0 < \alpha < 0.5$. For every adversarial client, $C'_i$, the attacker is able to add a trigger $\delta$ to every sample in its training set $x\in \mathcal{D}_i$. The goal of the attacker is to fool the central server generator to produce corrupted images that do not have medical research value.

\subsection{Defense Strategy}
\label{method:defence}
As introduced in Section~\ref{rw:defense}, the existing defense strategies focus on classification models, ranging from model level to data level. As data are not accessible in FL, model level defense is desired, where a model level detector is built to find the adversarial behavior and refrain it from training with others \citep{guo2021overview}, known as \textit{malicious detection}. Apart from detection, \textit{robust training} is another approach that refines training protocol~\cite{ozdayi2021defending}. To the best of our knowledge, \emph{no existing study} addresses the defense for FedGAN. In this section, we introduce an efficient and effective defense method, \ours{}.

\subsubsection{Defender’s capabilities} Let’s recall from our setting that a trusted server and more than half of the benign clients are part of our trusted FL pipeline. The benign server has access to the \textit{model parameters} and requests \textit{training loss}. Note that sharing training loss barely impacts data privacy. At the same time, the server cannot access the local data in any form given the privacy enforcement. Furthermore, the FedGAN server has control over the aggregation process, \eg, performs anomaly detection through locally updated information and assigns suspicious clients low weights to finally exclude them.

Our defense strategy is motivated by the observation that models with backdoor attacks tend to overfit the trigger rather than the actual image \citep{bagdasaryan2020backdoor}. Specifically, in GAN's training, the discriminator overfits the trigger and perfectly classifies fake and real images, while the generator does not receive effective feedback from the discriminator and then yields high loss and even diverges. Mathematically, as the discriminator overfits on the trigger, $\log{(1-D(G(z|y)))}$ in equation~\ref{eq:GenLoss} will ultimately over-shots and corrupt the performance of GAN. The locally corrupted GAN gradually contaminates other benign clients through the global aggregation. To this end, we propose to defend against backdoor attacks from the global level by leveraging malicious detection.

\subsubsection{\ours{}}
Our proposed defense strategy is based on two key observations:
\begin{itemize}
    \item The malicious clients with poisoning images can easily overfit discriminating the triggers, resulting in worse generator training performance.
    \item In backdoor attacks, the benign clients' model updates and loss patterns from a client in multiple iterations are consistent. 
\end{itemize}

\begin{algorithm*}[t]
    \caption{Establish iTree}
    \label{alg:itree}
    \textbf{Notations: }{ $\mathcal{L}$ represents the set of generator loss. $\phi$ is the sub-sampling size.}
	\begin{algorithmic}[1] 
	\Procedure{\textbf{A}. Establish-iTree}{$\mathcal{L}$} 
	            \If{$\mathcal{L} \leq \phi$}
	            \Comment{Construct the tree if reached the sub-sampling size.}
	            \State {\textbf{return} Node($\mathcal{L}$)}
	            \Else
                \State{$\mathcal{L}_l, \mathcal{L}_r \leftarrow$ split $\mathcal{L}$ according to certain attribute}
                \Comment{Randomly split point according to certain attribute}
                \State{\textbf{return} \Call{Establish-iTree}{$\mathcal{L}_l$} $\cup$ \Call{Establish-iTree}{$\mathcal{L}_r$}}
                \Comment{Recursively partition to form iTree}
                \EndIf
	\EndProcedure
	\end{algorithmic}
\end{algorithm*}

\noindent \textbf{Suspicious clients detection}
Based on the above two observations, the assumption that malicious clients are the minorities of the system and the FL setting (\ie, generators $G_i$ are shared and discriminator $D_i$'s information is kept locally) discussed earlier, we suggest performing anomaly detection using the generator loss for client $C_i$ at iteration $t$ specified as
\begin{equation}
    \ell_{G_i}^{(t)} = \mathbb{E}_{z\sim p_{\text{z}}(z)}[ \log{(1-D_i^{(t)}(G_i^{(t)}(z_i|y_i)))}].
\end{equation}
Then, Isolation Forests~\cite{liu2008isolation} is used for anomalous clients detection in every epoch on the generator losses $\mathcal{L}=\{\ell_{G_1},\dots, \ell_{G_N}\}$. $\mathcal{L}$ is then sent to the isolation forest for anomaly detection, which consists of two stages: fitting and predicting. The fitting stage is described in Algorithm~\ref{alg:itree}. It constructs isolation trees (iTrees) through recursively partitioning $\mathcal{L}$ (line 6-7) until reaching a sub-sampling size $\phi$ (line 3). This way, the potential outliers (points far away from inliers) are expected to be isolated with fewer partitions, which locates in the shallow leaves of iTrees as shown in Fig~\ref{fig:framework} (e). After constructing isolation forest that consists of a certain number of iTrees, it reaches the predicting stage, which first computes the single path length $h(\ell_{G_i}) = e + C(n)$ for every uploaded $\ell_{G_i}$, where $e$ stands for the number of edges from the root to $\ell_{G_i}$ in the iTree and $C(n) = 2H(n-1)-\frac{2(n-1)}{n}$. Ultimately, isolation forest computes the expectation score $Score(\ell_{G_i}) = 2^{-\frac{E(h(\ell_{G_i}))}{C(\phi)}}$. If $\ell_{G_i}$ has lesser length in most iTrees in the forest, $Score(\ell_{G_1})$ approaches 1, indicating higher possibility that $\ell_{G_1}$ is an outlier. 

To achieve a robust FL system without sacrificing privacy, the server in FedGAN only asks clients to report their loss along with the model parameters of the generator and perform outlier detection on the server-side. Again, in the FL setting for the medical applications, we assume all the clients' training protocol and reported value is regularized by the launched trustworthy software~\citep{pillutla2019robust}, thus the shared information is trusted. 

\noindent \textbf{Adaptive weights for robust aggregation}
As the initialization of FL training, we assign every client with an initial weight $ w_i = \frac{1}{|N|}$. Starting from epoch $m$ as a warmup, we activate the Isolation Forest \cite{liu2008isolation} on clients' losses of generator $\ell_{G_i}$ to red flag suspicious clients. Recall that there are less than half of malicious clients in our adversarial model. Thus, the valid detection algorithm should produce a set of potential malicious clients  $O$, where $|O| < \frac{1}{2}|N|$. We perform malicious detection per global iteration and keep track of the number of `malicious' red flags assigned to each client $C_i$ over the training process, denoting as counter $c_i$. In each global iteration, the adaptive aggregation weight of clients detected as an outlier will decay according to a decay constant $d \in (0,1)$ and the total time it has been detected $c_i$ by $w_i^{(t+1)} = w_i^{(t)\times d^{c_i}}$. Namely, if a client is more frequently detected as malicious, it receives a smaller aggregation weight. We further normalize the weights. The detailed algorithm is described in Algorithm~\ref{alg:cap}.

\section{Experiments}
This section presents experiment settings and results of backdoor attacks, \ours{}, state-of-the-art (SOTA) defense practices for FedGAN and FedGAN-assisted diagnoses based on two publicly available medical datasets. Section~\ref{exp:setting} introduces the details of both datasets and our FedGAN pipeline. Section~\ref{exp:attack} explores the effect of backdoor attack on both datasets with different types and sizes of triggers. Section~\ref{exp:defence} implements \ours{} and compare it with SOTA. Finally, section~\ref{exp:classification} trains a diagnostic model with different combinations of real data with vanilla, attack, and defense data to examine our FedGAN's performance in the medical scenario.

\subsection{Experimental Settings}
\label{exp:setting}
\subsubsection{Datasets} 
We experiment our FedGAN model on one RGB dataset and one black-white single-channel dataset.

\textbf{ISIC Dataset}
The International Skin Imaging Collaboration (ISIC) dataset \citep{codella2018skin} is a dataset for skin lesion classification widely used for proof of concept in medical image analysis. Our training set for all the clients of FedGAN contains 1800 Benign samples and 1497 Malignant samples. In our FedGAN, 824 randomly sampled images are assigned to each of the three clients and the last client gets 825 images. All images are resized to 256 $\times$ 256 before being fed into the FedGAN. We present sample ISIC images in Fig.~\ref{fig:isicAttack} (a).

\textbf{Chest X-Ray Dataset (Pneumonia)} ChestX~\citep{kermany2018identifying} contains 5856 gray-scale images. In our FedGAN, 1464 images are randomly sampled for each client. All images are also resized to the size of 256 $\times$ 256 as the ISIC dataset. The samples are present in Fig.~\ref{fig:chestAttack}(a).

\subsubsection{Generative Adversarial Networks:} Our conditional FedGAN follows the recipe of DCGAN~\citep{radford2015unsupervised}, one of the most widely used conditional GAN frameworks, as our basic network.  We replace the generator architecture of DCGAN with that of StyleGAN2-ADA \citep{karras2020training}, given its generator produces images with high qualities in the majority of datasets and has great potential to generate high-resolution medical images for clinical research~\citep{woodland2022evaluating}. It is worth noting that the attack and defense strategies although demonstrated with our selected architecture, they have the potential to apply to other GAN-based generative models. In all training scenarios, we adopt the Adam optimizer with a learning rate of $2 \times 10^{-4}$ for both the generator and the discriminator. The batch size is set to 32 as per the limit of a 32GB Tesla V100 GPU.

\subsubsection{Federated Learning}
\label{exp:fl}
\begin{figure*}[ht]
    \centering
    \captionsetup[subfloat]{labelformat=empty}
    \subfloat[(a) Original ISIC Images]{
    \includegraphics[width=0.078\textwidth, height=0.078\textwidth]{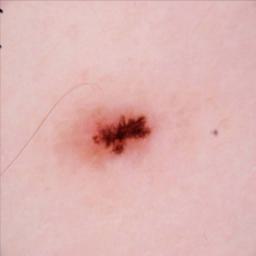}
    \includegraphics[width=0.078\textwidth, height=0.078\textwidth]{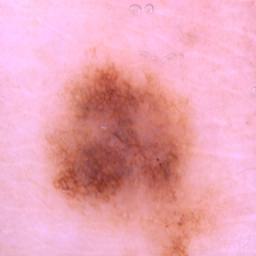}
    \includegraphics[width=0.078\textwidth, height=0.078\textwidth]{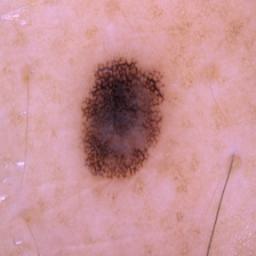}
    \includegraphics[width=0.078\textwidth, height=0.078\textwidth]{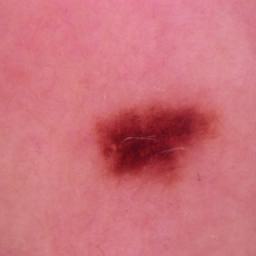}
    \includegraphics[width=0.078\textwidth, height=0.078\textwidth]{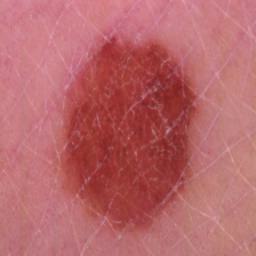}
    \includegraphics[width=0.078\textwidth, height=0.078\textwidth]{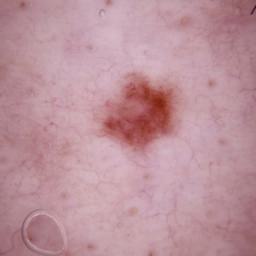}
    \includegraphics[width=0.078\textwidth, height=0.078\textwidth]{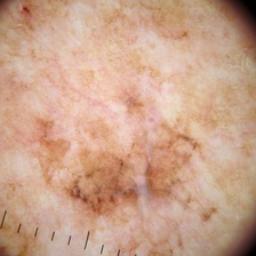}
    \includegraphics[width=0.078\textwidth, height=0.078\textwidth]{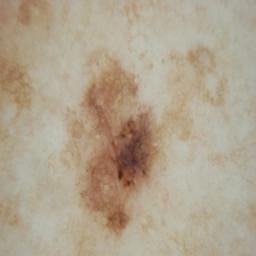}
    \includegraphics[width=0.078\textwidth, height=0.078\textwidth]{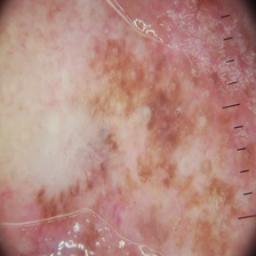}
    \includegraphics[width=0.078\textwidth, height=0.078\textwidth]{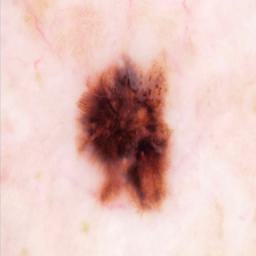}
    \includegraphics[width=0.078\textwidth, height=0.078\textwidth]{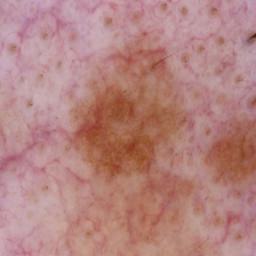}
    \includegraphics[width=0.078\textwidth, height=0.078\textwidth]{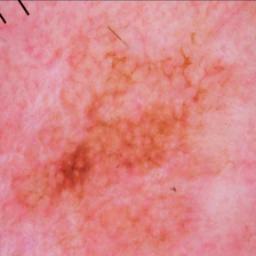}
}
    
    \subfloat[(b) Vanilla FedGAN]{
    \includegraphics[width=0.078\textwidth, height=0.078\textwidth]{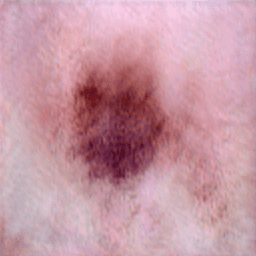}
    \includegraphics[width=0.078\textwidth, height=0.078\textwidth]{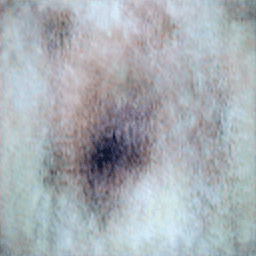}
    \includegraphics[width=0.078\textwidth, height=0.078\textwidth]{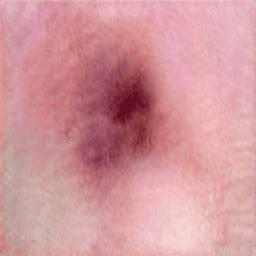}
    \includegraphics[width=0.078\textwidth, height=0.078\textwidth]{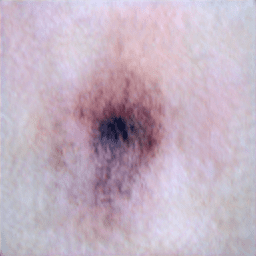}
    \includegraphics[width=0.078\textwidth, height=0.078\textwidth]{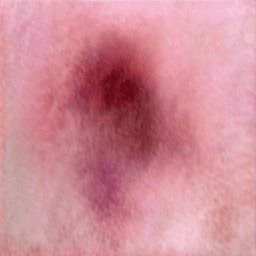}
    \includegraphics[width=0.078\textwidth, height=0.078\textwidth]{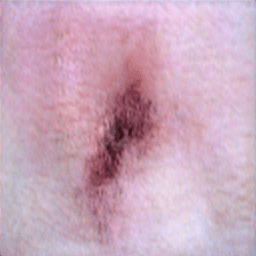}
    \includegraphics[width=0.078\textwidth, height=0.078\textwidth]{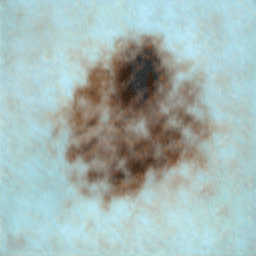}
    \includegraphics[width=0.078\textwidth, height=0.078\textwidth]{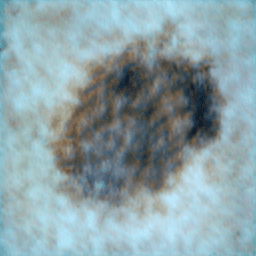}
    \includegraphics[width=0.078\textwidth, height=0.078\textwidth]{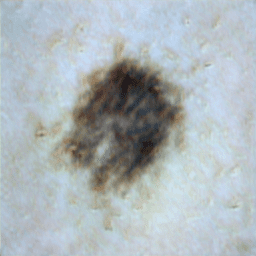}
    \includegraphics[width=0.078\textwidth, height=0.078\textwidth]{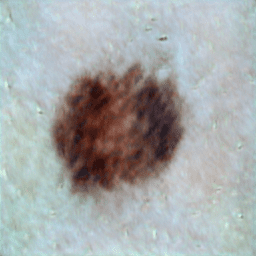}
    \includegraphics[width=0.078\textwidth, height=0.078\textwidth]{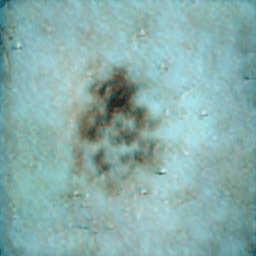}
    \includegraphics[width=0.078\textwidth, height=0.078\textwidth]{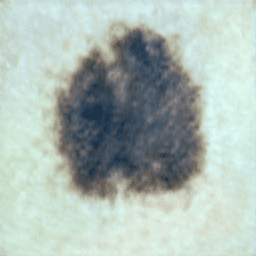}
}

\subfloat[(c) Attack with Poison Size of 16x16]{
    \includegraphics[width=0.078\textwidth, height=0.078\textwidth]{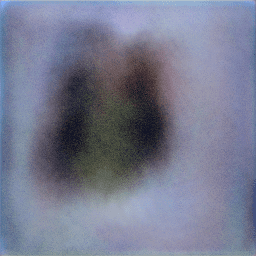}
    \includegraphics[width=0.078\textwidth, height=0.078\textwidth]{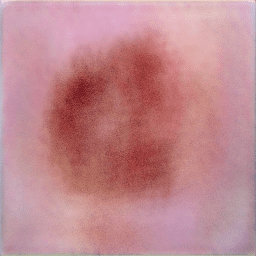}
    \includegraphics[width=0.078\textwidth, height=0.078\textwidth]{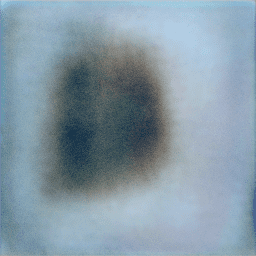}
    \includegraphics[width=0.078\textwidth, height=0.078\textwidth]{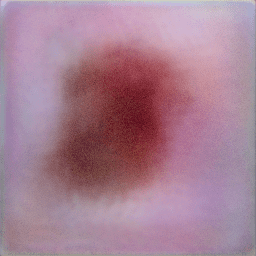}
    \includegraphics[width=0.078\textwidth, height=0.078\textwidth]{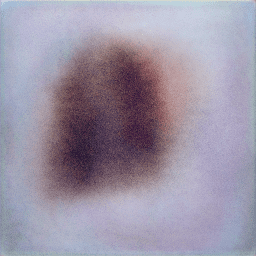}
    \includegraphics[width=0.078\textwidth, height=0.078\textwidth]{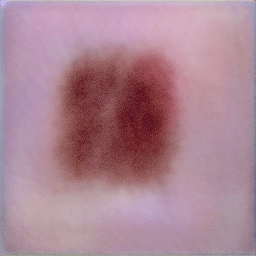}
    \includegraphics[width=0.078\textwidth, height=0.078\textwidth]{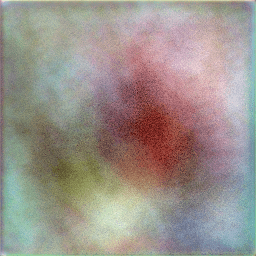}
    \includegraphics[width=0.078\textwidth, height=0.078\textwidth]{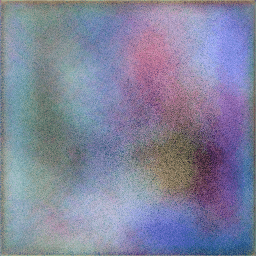}
    \includegraphics[width=0.078\textwidth, height=0.078\textwidth]{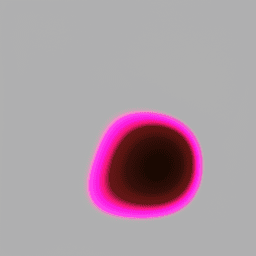}
    \includegraphics[width=0.078\textwidth, height=0.078\textwidth]{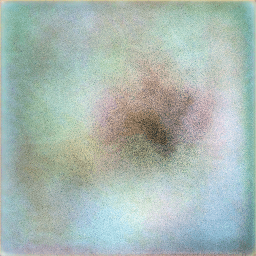}
    \includegraphics[width=0.078\textwidth, height=0.078\textwidth]{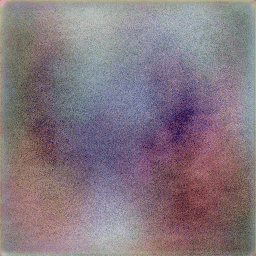}
    \includegraphics[width=0.078\textwidth, height=0.078\textwidth]{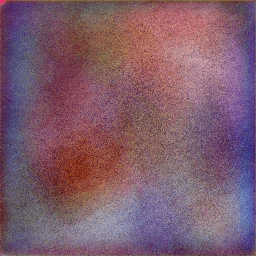}
}

\subfloat[(d) Attack with Poison Size of 32x32]{
    \includegraphics[width=0.078\textwidth, height=0.078\textwidth]{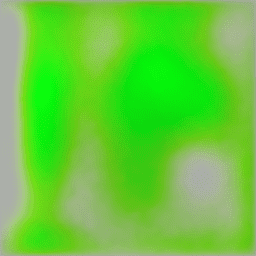}
    \includegraphics[width=0.078\textwidth, height=0.078\textwidth]{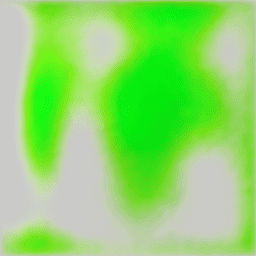}
    \includegraphics[width=0.078\textwidth, height=0.078\textwidth]{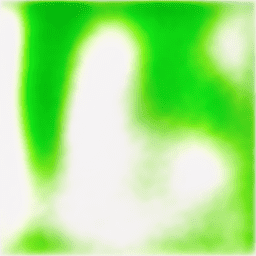}
    \includegraphics[width=0.078\textwidth, height=0.078\textwidth]{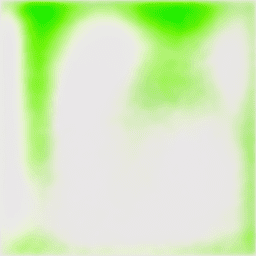}
    \includegraphics[width=0.078\textwidth, height=0.078\textwidth]{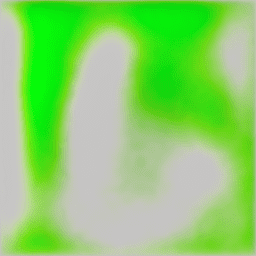}
    \includegraphics[width=0.078\textwidth, height=0.078\textwidth]{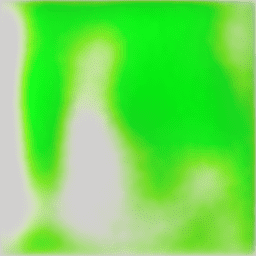}
    \includegraphics[width=0.078\textwidth, height=0.078\textwidth]{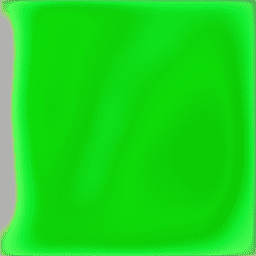}
    \includegraphics[width=0.078\textwidth, height=0.078\textwidth]{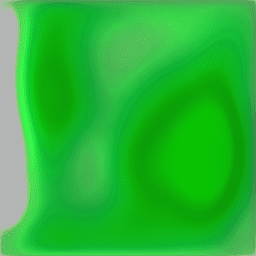}
    \includegraphics[width=0.078\textwidth, height=0.078\textwidth]{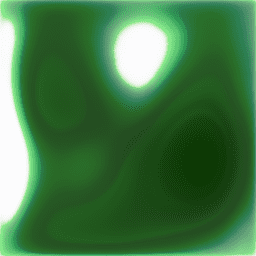}
    \includegraphics[width=0.078\textwidth, height=0.078\textwidth]{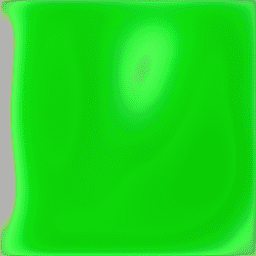}
    \includegraphics[width=0.078\textwidth, height=0.078\textwidth]{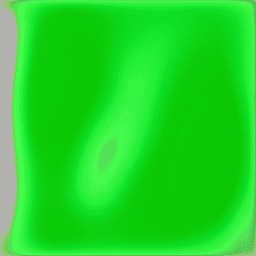}
    \includegraphics[width=0.078\textwidth, height=0.078\textwidth]{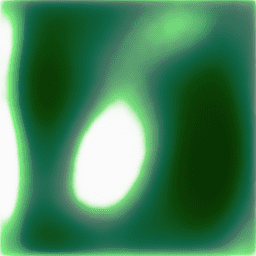}
}

\subfloat[(e) Attack with Poison Size of 64x64]{
    \includegraphics[width=0.078\textwidth, height=0.078\textwidth]{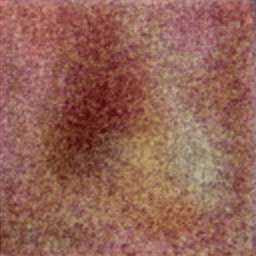}
    \includegraphics[width=0.078\textwidth, height=0.078\textwidth]{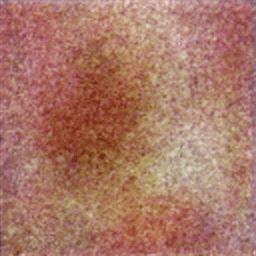}
    \includegraphics[width=0.078\textwidth, height=0.078\textwidth]{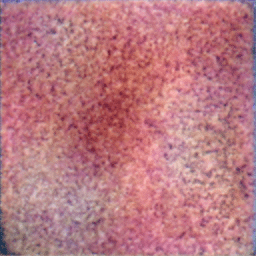}
    \includegraphics[width=0.078\textwidth, height=0.078\textwidth]{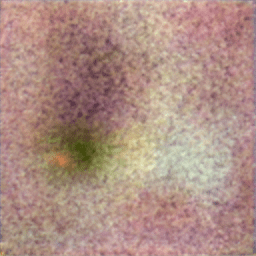}
    \includegraphics[width=0.078\textwidth, height=0.078\textwidth]{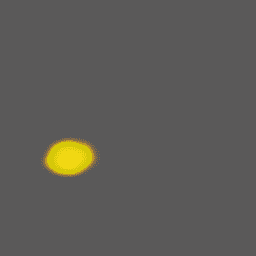}
    \includegraphics[width=0.078\textwidth, height=0.078\textwidth]{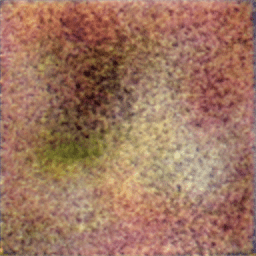}
    \includegraphics[width=0.078\textwidth, height=0.078\textwidth]{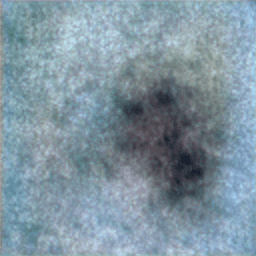}
    \includegraphics[width=0.078\textwidth, height=0.078\textwidth]{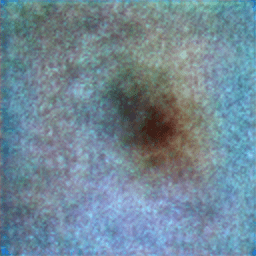}
    \includegraphics[width=0.078\textwidth, height=0.078\textwidth]{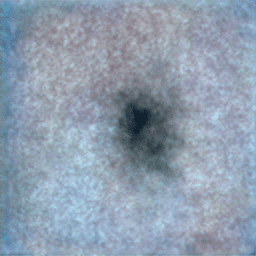}
    \includegraphics[width=0.078\textwidth, height=0.078\textwidth]{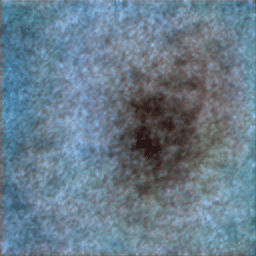}
    \includegraphics[width=0.078\textwidth, height=0.078\textwidth]{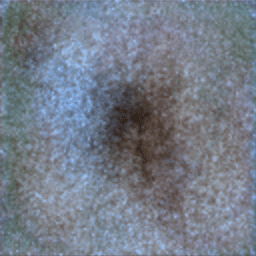}
    \includegraphics[width=0.078\textwidth, height=0.078\textwidth]{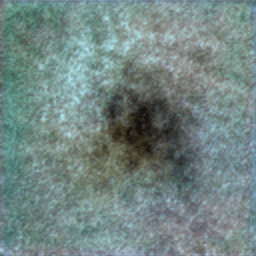}
}
    \caption{Visualization on ISIC Attacking}
    \label{fig:isicAttack}
\end{figure*}

\begin{figure*}[ht]
    \centering
    \captionsetup[subfloat]{labelformat=empty}
    \subfloat[(a) Original ChestX Images]{
    \includegraphics[width=0.078\textwidth, height=0.078\textwidth]{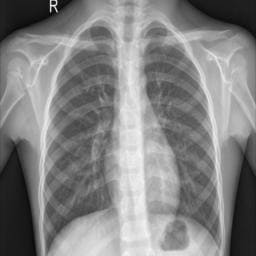}
    \includegraphics[width=0.078\textwidth, height=0.078\textwidth]{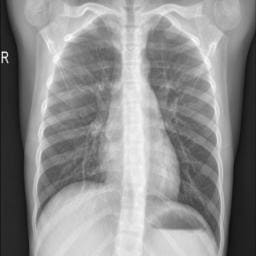}
    \includegraphics[width=0.078\textwidth, height=0.078\textwidth]{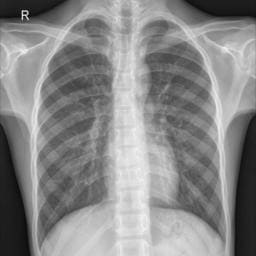}
    \includegraphics[width=0.078\textwidth, height=0.078\textwidth]{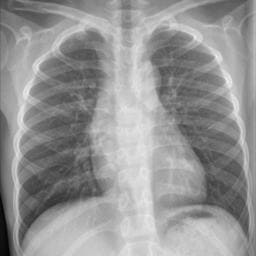}
    \includegraphics[width=0.078\textwidth, height=0.078\textwidth]{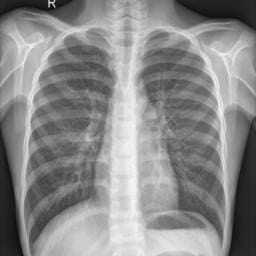}
    \includegraphics[width=0.078\textwidth, height=0.078\textwidth]{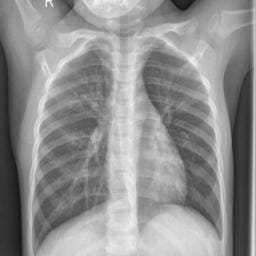}
    \includegraphics[width=0.078\textwidth, height=0.078\textwidth]{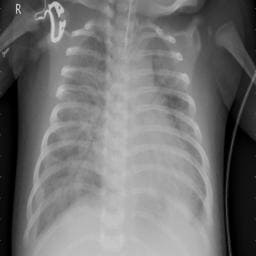}
    \includegraphics[width=0.078\textwidth, height=0.078\textwidth]{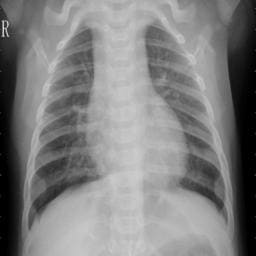}
    \includegraphics[width=0.078\textwidth, height=0.078\textwidth]{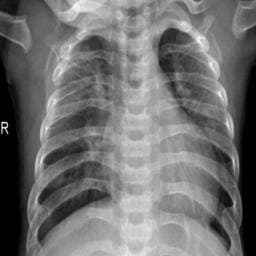}
    \includegraphics[width=0.078\textwidth, height=0.078\textwidth]{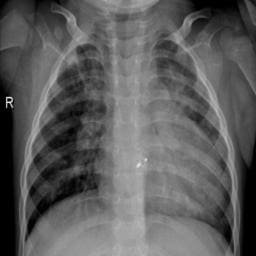}
    \includegraphics[width=0.078\textwidth, height=0.078\textwidth]{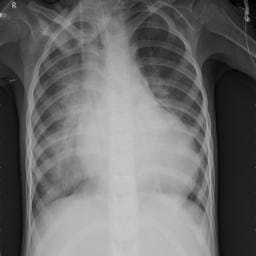}
    \includegraphics[width=0.078\textwidth, height=0.078\textwidth]{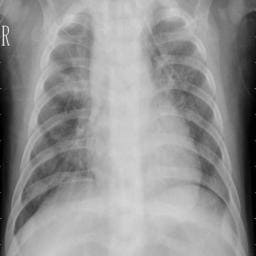}
    }
    
    \subfloat[(b) Vanilla FedGAN]{
    \includegraphics[width=0.078\textwidth, height=0.078\textwidth]{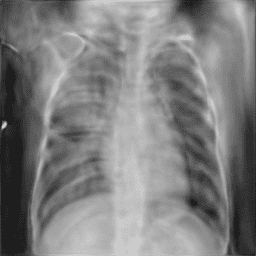}
    \includegraphics[width=0.078\textwidth, height=0.078\textwidth]{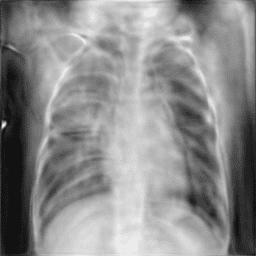}
    \includegraphics[width=0.078\textwidth, height=0.078\textwidth]{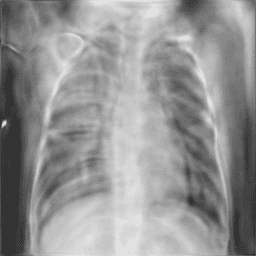}
    \includegraphics[width=0.078\textwidth, height=0.078\textwidth]{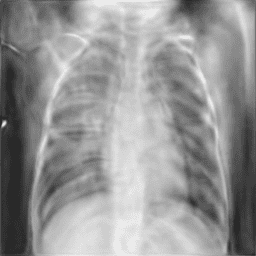}
    \includegraphics[width=0.078\textwidth, height=0.078\textwidth]{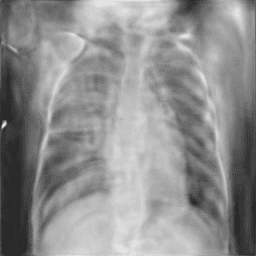}
    \includegraphics[width=0.078\textwidth, height=0.078\textwidth]{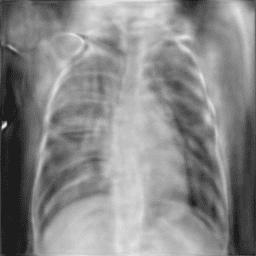}
    \includegraphics[width=0.078\textwidth, height=0.078\textwidth]{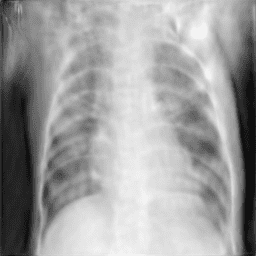}
    \includegraphics[width=0.078\textwidth, height=0.078\textwidth]{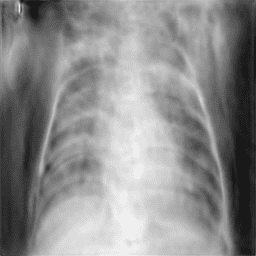}
    \includegraphics[width=0.078\textwidth, height=0.078\textwidth]{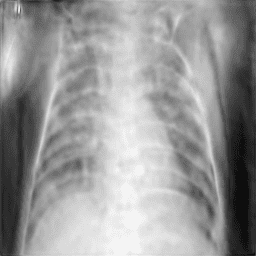}
    \includegraphics[width=0.078\textwidth, height=0.078\textwidth]{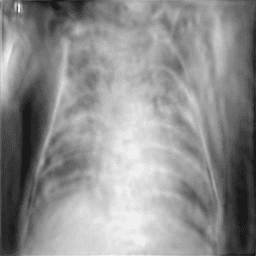}
    \includegraphics[width=0.078\textwidth, height=0.078\textwidth]{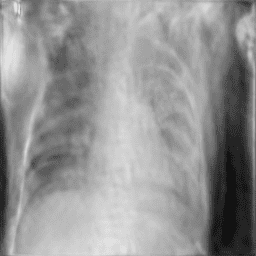}
    \includegraphics[width=0.078\textwidth, height=0.078\textwidth]{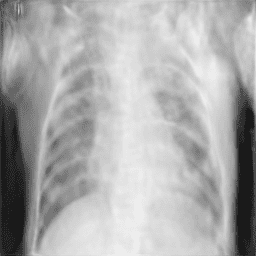}
}

\subfloat[(c) Attack with Poison Size of 16x16]{
    \includegraphics[width=0.078\textwidth, height=0.078\textwidth]{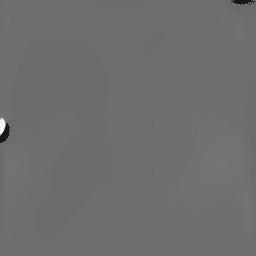}
    \includegraphics[width=0.078\textwidth, height=0.078\textwidth]{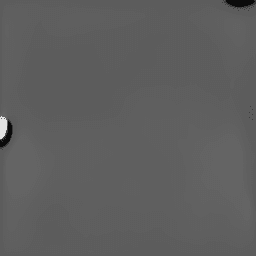}
    \includegraphics[width=0.078\textwidth, height=0.078\textwidth]{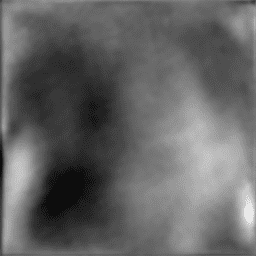}
    \includegraphics[width=0.078\textwidth, height=0.078\textwidth]{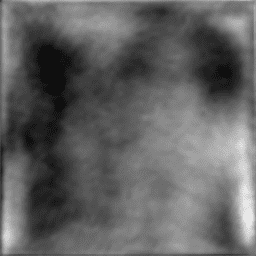}
    \includegraphics[width=0.078\textwidth, height=0.078\textwidth]{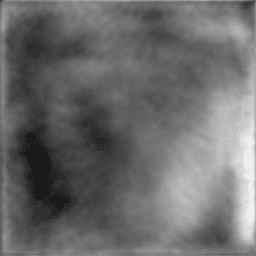}
    \includegraphics[width=0.078\textwidth, height=0.078\textwidth]{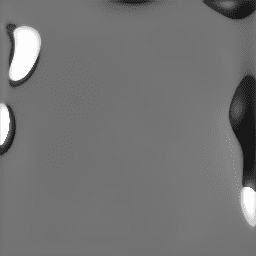}
    \includegraphics[width=0.078\textwidth, height=0.078\textwidth]{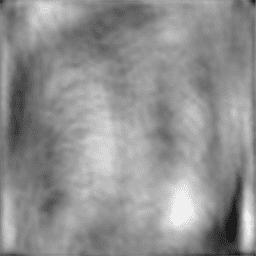}
    \includegraphics[width=0.078\textwidth, height=0.078\textwidth]{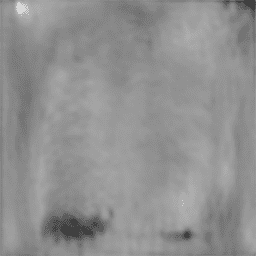}
    \includegraphics[width=0.078\textwidth, height=0.078\textwidth]{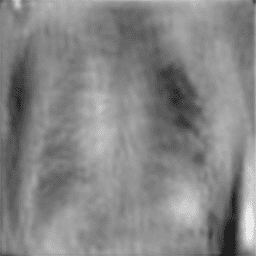}
    \includegraphics[width=0.078\textwidth, height=0.078\textwidth]{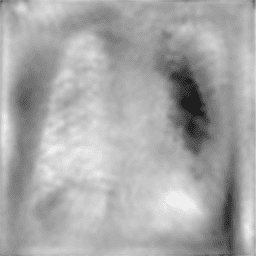}
    \includegraphics[width=0.078\textwidth, height=0.078\textwidth]{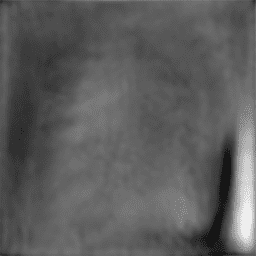}
    \includegraphics[width=0.078\textwidth, height=0.078\textwidth]{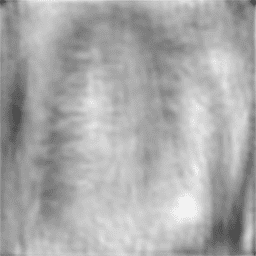}
}

\subfloat[(d) Attack with Poison Size of 32x32]{
    \includegraphics[width=0.078\textwidth, height=0.078\textwidth]{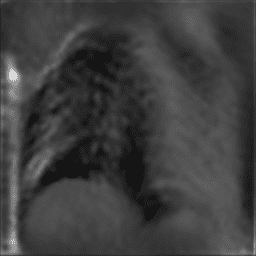}
    \includegraphics[width=0.078\textwidth, height=0.078\textwidth]{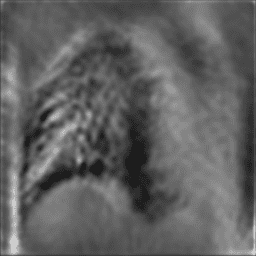}
    \includegraphics[width=0.078\textwidth, height=0.078\textwidth]{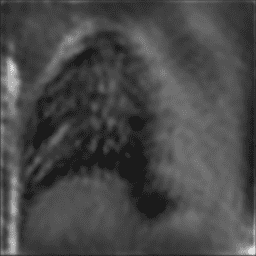}
    \includegraphics[width=0.078\textwidth, height=0.078\textwidth]{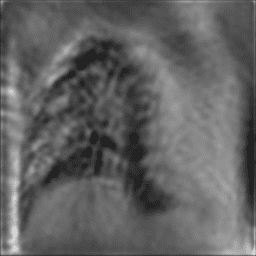}
    \includegraphics[width=0.078\textwidth, height=0.078\textwidth]{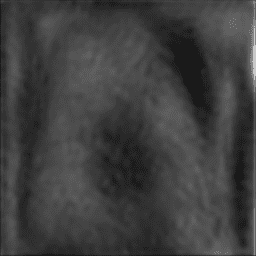}
    \includegraphics[width=0.078\textwidth, height=0.078\textwidth]{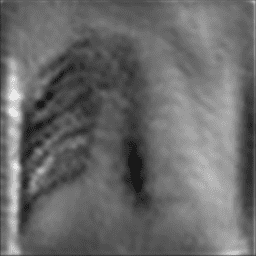}
    \includegraphics[width=0.078\textwidth, height=0.078\textwidth]{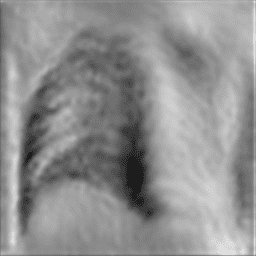}
    \includegraphics[width=0.078\textwidth, height=0.078\textwidth]{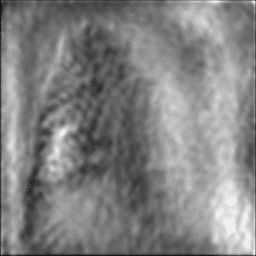}
    \includegraphics[width=0.078\textwidth, height=0.078\textwidth]{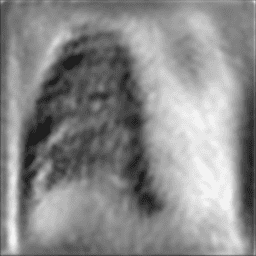}
    \includegraphics[width=0.078\textwidth, height=0.078\textwidth]{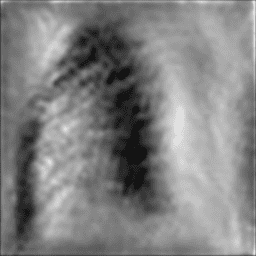}
    \includegraphics[width=0.078\textwidth, height=0.078\textwidth]{figs/chestAttack/ak32/pneumonia/fake_4.png}
    \includegraphics[width=0.078\textwidth, height=0.078\textwidth]{figs/chestAttack/ak32/pneumonia/fake_5.png}
}

\subfloat[(e) Attack with Poison Size of 64x64]{
    \includegraphics[width=0.078\textwidth, height=0.078\textwidth]{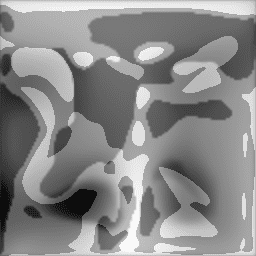}
    \includegraphics[width=0.078\textwidth, height=0.078\textwidth]{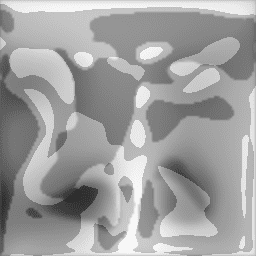}
    \includegraphics[width=0.078\textwidth, height=0.078\textwidth]{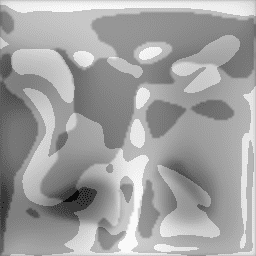}
    \includegraphics[width=0.078\textwidth, height=0.078\textwidth]{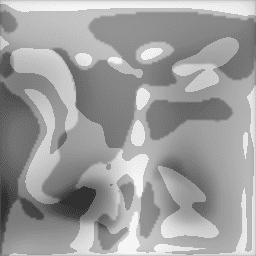}
    \includegraphics[width=0.078\textwidth, height=0.078\textwidth]{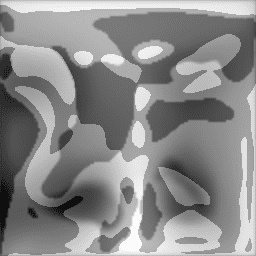}
    \includegraphics[width=0.078\textwidth, height=0.078\textwidth]{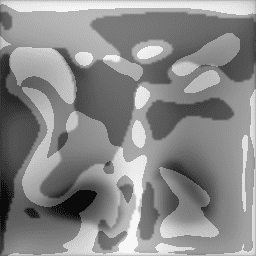}
    \includegraphics[width=0.078\textwidth, height=0.078\textwidth]{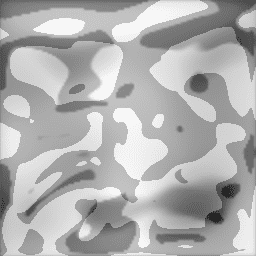}
    \includegraphics[width=0.078\textwidth, height=0.078\textwidth]{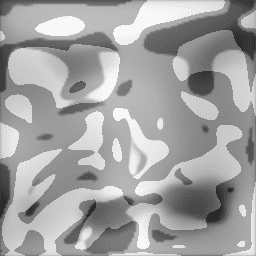}
    \includegraphics[width=0.078\textwidth, height=0.078\textwidth]{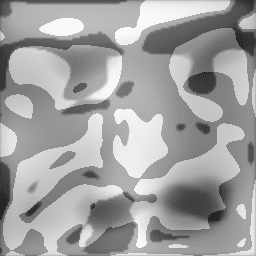}
    \includegraphics[width=0.078\textwidth, height=0.078\textwidth]{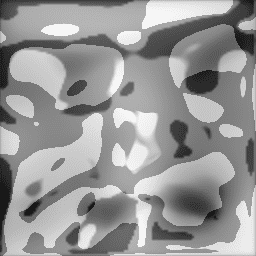}
    \includegraphics[width=0.078\textwidth, height=0.078\textwidth]{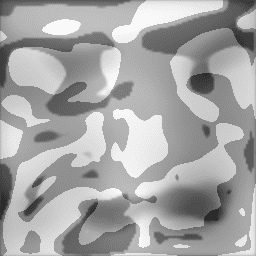}
    \includegraphics[width=0.078\textwidth, height=0.078\textwidth]{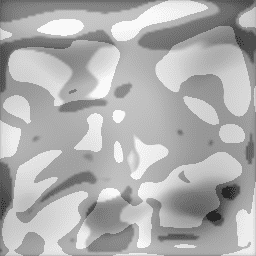}
}
    \caption{Visualization on ChestX Attacking}
    \label{fig:chestAttack}
\end{figure*}

Considering the total available sample size and the limitation of our GPU and hardware memory, we run FedGAN on four clients where each client is trained on randomly equally sampled images from the datasets described above. This study serves as a first step towards understanding backdoor attacks and exploring defense strategies in FedGAN training for medical image synthesis. We contend scale of simulation is still significant and can easily be generalized to a larger scale. We update the local generator parameters to the global server every local epoch and train the FedGAN with 200 global epochs using FedAvg~\cite{mcmahan2017communication}. The synthetic medical images with vanilla FedGAN (no attack and defense induced) are presented in Fig.~\ref{fig:isicAttack} (b) for ISIC dataset and Fig.~\ref{fig:chestAttack} (b) for ChestX dataset.

\subsection{Backdoor Attack on FedGAN}
\label{exp:attack}

\begin{table*}[ht]
\centering
\setlength\tabcolsep{5pt}
\caption{Quantitative Comparison for Attack. $\downarrow$ indicates the smaller the better. }
\label{tab:isicAttack}
\begin{tabular}{@{}ccccc@{}}
\toprule
\multirow{3}{*}{Settings} &
\multicolumn{4}{c}{ISIC} \\
\cmidrule(l){2-5}
&
\multirow{1}{*}{Vanilla GAN}&
\multicolumn{3}{c}{Attack}\\
\cmidrule(l){3-5}
& & 16x16 & 32x32 & 64x64\\ 
\midrule
LPIPS (Benign) $\downarrow$ &     0.6735  &   0.7005 & 1.0373 & 0.7187 \\
LPIPS (Malignant) $\downarrow$ &     0.7354    &   0.9248 & 1.0337 & 0.7984 \\
\bottomrule
\end{tabular}
\end{table*}

\begin{table*}[ht]
\centering
\setlength\tabcolsep{5pt}
\caption{Quantitative Comparison for Attack. $\downarrow$ indicates the smaller the better. }
\label{tab:chestAttack}
\begin{tabular}{@{}ccccc@{}}
\toprule
\multirow{3}{*}{Settings} &
\multicolumn{4}{c}{ChestX} \\
\cmidrule(l){2-5}
&
\multirow{1}{*}{Vanilla GAN}&
\multicolumn{3}{c}{Attack}\\ 
\cmidrule(l){3-5}
 & & 16x16 & 32x32 & 64x64 \\ 
\midrule
LPIPS (Normal) $\downarrow$ & 0.3748 & 0.7479 & 0.6193 & 0.6042 \\
LPIPS (Pneumonia) $\downarrow$ & 0.4158 & 0.5968 & 0.6327 & 0.6165 \\
\bottomrule
\end{tabular}
\end{table*}

\subsubsection{Backdoor Details}
As mentioned in Section.~\ref{exp:fl}, four clients are simulating four medical institutions in our FedGAN setting and each client is trained on equally sampled data from the original training dataset. Among the four clients, one is randomly selected as the malicious client. The three benign clients are trained with normal images, while the malicious client is trained with poisoned images. As for the poisoned images, different triggers are applied for ISIC and ChestX datasets as detailed below based on their modalities.

\noindent\textbf{Backdoor on ISIC}
We apply the trigger strategy proposed by~\cite{saha2020hidden}, which has shown to be effective for backdoor attacks in classification tasks. Specifically, we adopt a random matrix of colors that has a different pattern from the actual image with sizes ranging from 16 $\times$ 16 (only 0.39\% of the size of the original image) to 64 $\times$ 64. The same trigger is pasted onto the bottom right of all the training images in the malicious client before launching FedGAN training. The examples of poisoned images are shown in Fig.~\ref{fig:framework}, which are fed into the discriminator $D$ of malicious clients. 

\noindent\textbf{Backdoor on ChestX}
ChestX is a gray-scale image dataset and it will be inappropriate to use the same trigger as the ISIC dataset. In this case, we attempted to use a white square as the trigger. All the rest settings are the same as above, including the sizes of the trigger starting from $16 \time 16$ to $64 \times 64$ and its location is on the bottom right of the image. 

\subsubsection{Metrics} In addition to the qualitative evaluation of the fidelity of synthetic images, we have also included the Learned Perceptual Image Patch Similarity (LPIPS) to quantitatively evaluate the them~\cite{zhang2018unreasonable}. LPIPS assesses the perceptual similarity between two pictures. In essence, LPIPS determines how comparable two picture patches' activations are for a given network. In our experiments, we use AlexNet as the neural network. This measurement has been demonstrated to closely reflect human perception~\citep{wang2020deep}. Image patches with a low LPIPS score are perceptually similar. We have also run the classification experiments in Section~\ref{exp:classification} to measure the influence of attack directly on training Deep Neural Networks.

\subsubsection{Attack Result and Analysis} 
\noindent \textbf{Visualization Results}  The visualization of the original images for both datasets are shown in Fig~\ref{fig:isicAttack}(a) and Fig~\ref{fig:chestAttack}(a). The generated images are shown in Fig.~\ref{fig:isicAttack}(b) and Fig~\ref{fig:chestAttack}(b). In both Fig~\ref{fig:isicAttack} and Fig~\ref{fig:chestAttack}, (c), (d) and (e) presents the attacked images with poison size of 16x16, 32x32 and 64x64 individually for both datasets. The virtualization plot shows that the appearances of the attacked images are significantly different from those generated images in row (b), which means that essential disease-related information is lost and the images cannot be used for diagnostic reasons. Furthermore, we find that the strongest assault, which completely ruins FedGAN performance, has a poison size of 32x32 for the ISIC dataset and 64x64 for the ChestX dataset. As the discriminator in our FedGAN is consistently maintained locally, we anticipate that the poisoned data will first directly impact the performance of the discriminator, yielding a sub-optimal discriminator. This subsequently affects the generator through adversarial training, which, in turn, contaminates other FL participants during server aggregation. Consequently, the size of the trigger has a limited direct impact on the strength of the attack, as the poisoned dataset only indirectly affects the entire system. However, we perform the ablation study for trigger sizes to demonstrate the generalizability of our attacks and defenses.

\noindent \textbf{Quantifying Image Fidelity} For most attacked images, we are able to observe a significant contrast and distortion from the benign images. In both datasets, we also report LPIPS values. The smaller the LPIPS, the more similar the two images. The quantitative metrics match what we observed, where the attacked images have higher LPIPS values than those of vanilla images, indicating that there is less similarity between the generated images with the original datasets than the vanilla GAN generated images. As shown in Table~\ref{tab:isicAttack}
and Table~\ref{tab:chestAttack}, the LPIPS significantly increases for both datasets. The 32x32 patch achieves the strongest attack, in which the LPIPS score increases 54\% compared to the Vanilla FedGAN. For the ChestX dataset, all LPIPS score for normal samples almost doubles compared to the Vanilla GAN, indicating the large gap in the similarity between the original images and the generated images.

\subsection{Defense Result and Analysis}
\label{exp:defence}
When the malicious clients train on contaminated data in the attack outlined in Section~\ref{exp:attack}, the discriminator soon overfits the trigger and causes the entire FedGAN model to suffer from training instability. Here, we try to carry out \ours{} to counter this threat and compare it with several alternatives.

\subsubsection{Baseline Defense Strategies}
\label{exp:moredf}
As summarized in Section~\ref{rw:defense}, state-of-the-art defense practices are made up of data level and model level. Data level defenses include adding trigger blockers, eliminating backdoors from the training data, and data argumentation. Due to the privacy regulations of FL, FedGAN is unable to record the clients' training data; as a result, the deleting trigger strategy is not applicable. Also, the existing defense strategies focus on classification models. 

Among the model level defense strategies, the majority of them also fail under the FedGAN pipeline. For example, \emph{the diagnostic approach} requires a certain pre-trained meta-classifier~\citep{kolouri2020universal, xu2021detecting}, which is inappropriate for GAN's scenario, where the discriminator and the generator need to train against each other at the beginning. \emph{The unlearning approach} takes two steps, first, the infected data need to be detected, and then the model will unlearn such data~\citep{li2021anti, zeng2021adversarial}. This strategy works well for classification tasks given its reversible objective function, but it again does not fit the scenario of GAN, where both the discriminator and the generator train against each other through the minmax loss function as specified in equation~\ref{eq:cganloss1}, which cannot be easily reversed as the classification loss forms. \emph{The trigger synthesis approach} requires the defender to access data information~\citep{wang2019neural, guo2020towards}. Additionally, because the FedGAN diverged before the trigger was synthesized in our example, the synthesizing trigger practice does not suit the FedGAN discipline.

In the end, we consider the comparison with the data argumentation technique and the model reconstruction strategy that can be compatible with the scenario of FedGAN. We deep dive into both methods in detail below and perform empirical experiments on both methods:

\noindent\textbf{Data Augmentation} To the best of our knowledge, ~\cite{li2020rethinking} first draw the conclusion that the effect of backdoor attack depends heavily on the position and the appearance of triggers. They proposed the spatial transformation-based defense, which modifies the location of the trigger on images, including horizontal flipping and padding after shrinking on the original image. \cite{qiu2021deepsweep,borgnia2021strong, zeng2021rethinking} further performed more comprehensive data augmentations and validated the efficacy of them in centralized classification models. In our setting, we apply two kinds of data augmentation. Given the backdoor attack is sensitive to the location of the trigger as indicated by~\cite{li2020rethinking}, we perform the random horizontal flipping on inputs before being fed into the FedGAN in the first experiment. In the second experiment, random rotation is applied. Each image will be randomly rotated within the range of $(-90, 90)$ and then fed into the FedGAN.

\noindent\textbf{Model Reconstruction} Model Reconstruction, also known as fine-tuning, aims to retrain the infected DL model with a small amount of benign data to alleviate the effect of backdoor attack~\citep{liu2017neural, li2022backdoor}. This strategy is based on the theory of the catastrophic forgetting property of the deep neural networks~\citep{kirkpatrick2017overcoming}. Since in FL, each client owns a specific quantity of benign data, thus model reconstruction matches the FedGAN setting. In order to simulate the scenarios in the reality, we reconstruct the model with a dataset containing 500 real images per class after receiving the trained generator model from the FedGAN server. In this experiment, we fine-tune the innocuous samples with 200 epochs and examine their performance.

\noindent\textbf{Robust Aggregation} In FL, the defense against backdoor attacks has recently incorporated the concept of robust aggregation, as demonstrated by techniques such as adjusting the ``learning rate" at the server level during the aggregation process~\citep{ozdayi2021defending,pillutla2019robust}. This approach assumes that the updated gradients from both malicious and benign clients exhibit element-wise sign differences. Consequently, a~\textit{threshold} is introduced at the server side, where the loss is maximized (e.g., a process of unlearning) if the sum of the signs falls below this~\textit{threshold}. The server utilizes a \textit{server learning rate} to control the step and direction during the aggregation process in order to maximize the loss. Specifically, the \textit{server learning rate} is multiplied by -1 for those dimensions where the sum of the signs is below the \textit{threshold}.

\subsubsection{Implementation Details of \ours{}} 
\begin{figure*}[ht]
    \centering
    \captionsetup[subfloat]{labelformat=empty}
    \subfloat[\normalsize (a) \ours{} (ours)]{
    \includegraphics[width=0.99\textwidth]{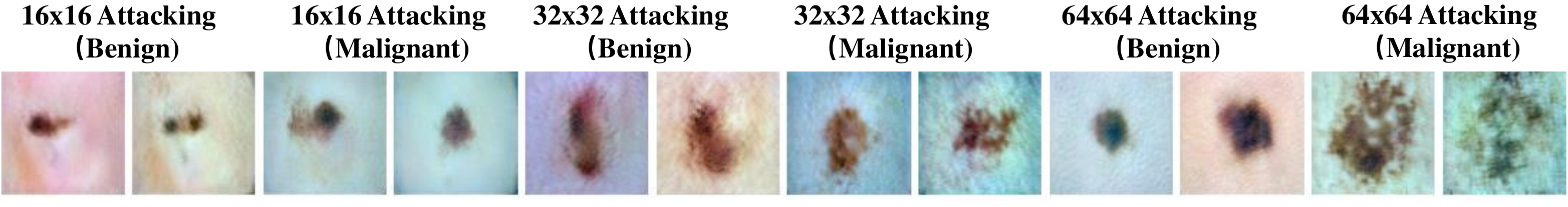}
}
    
    \subfloat[\normalsize (b) Model Reconstruction]{
    \includegraphics[width=0.078\textwidth, height=0.078\textwidth]{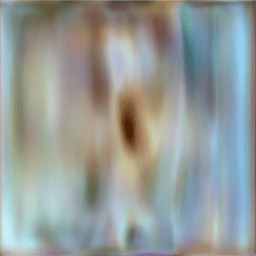}
    \includegraphics[width=0.078\textwidth, height=0.078\textwidth]{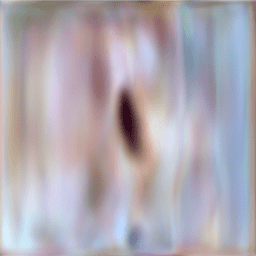}
     \includegraphics[width=0.078\textwidth, height=0.078\textwidth]{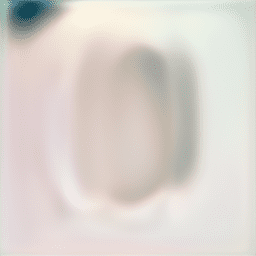}
    \includegraphics[width=0.078\textwidth, height=0.078\textwidth]{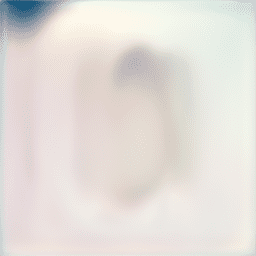}
    
    \includegraphics[width=0.078\textwidth, height=0.078\textwidth]{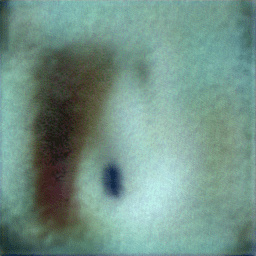}
    \includegraphics[width=0.078\textwidth, height=0.078\textwidth]{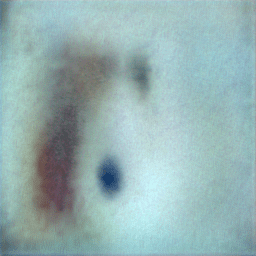}
     \includegraphics[width=0.078\textwidth, height=0.078\textwidth]{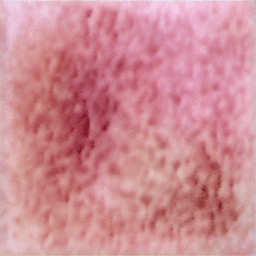}
    \includegraphics[width=0.078\textwidth, height=0.078\textwidth]{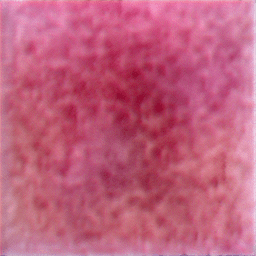}
    
    \includegraphics[width=0.078\textwidth, height=0.078\textwidth]{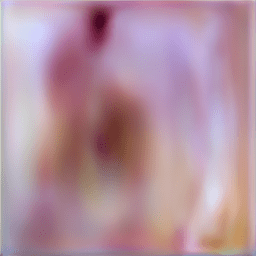}
    \includegraphics[width=0.078\textwidth, height=0.078\textwidth]{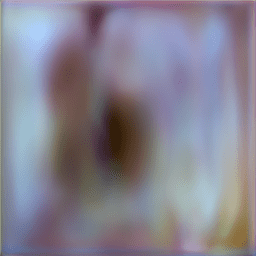}
     \includegraphics[width=0.078\textwidth, height=0.078\textwidth]{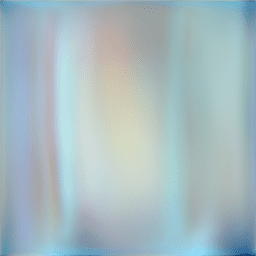}
    \includegraphics[width=0.078\textwidth, height=0.078\textwidth]{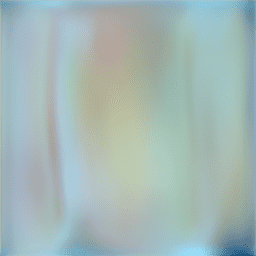}
}

\subfloat[\normalsize (c) Data Augmentation with Horizontal Flipping and Random Rotation]{
    \includegraphics[width=0.078\textwidth, height=0.078\textwidth]{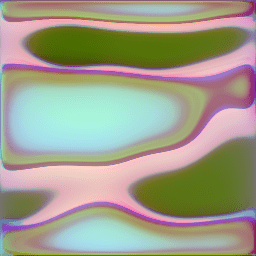}
    \includegraphics[width=0.078\textwidth, height=0.078\textwidth]{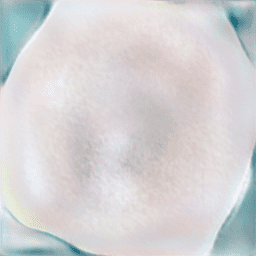}
     \includegraphics[width=0.078\textwidth, height=0.078\textwidth]{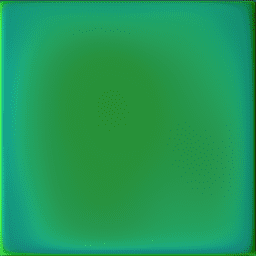}
    \includegraphics[width=0.078\textwidth, height=0.078\textwidth]{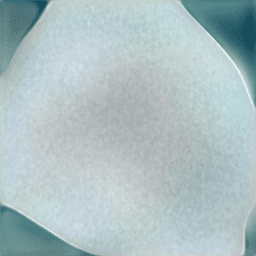}
    
    \includegraphics[width=0.078\textwidth, height=0.078\textwidth]{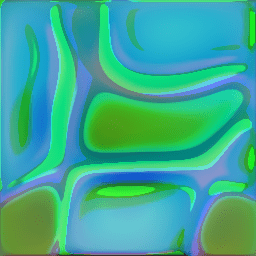}
    \includegraphics[width=0.078\textwidth, height=0.078\textwidth]{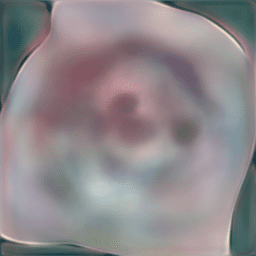}
     \includegraphics[width=0.078\textwidth, height=0.078\textwidth]{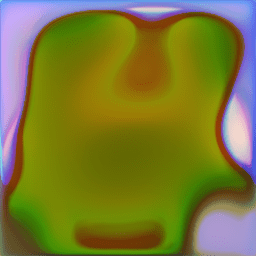}
    \includegraphics[width=0.078\textwidth, height=0.078\textwidth]{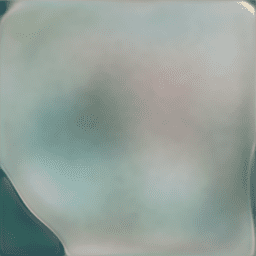}
    
    \includegraphics[width=0.078\textwidth, height=0.078\textwidth]{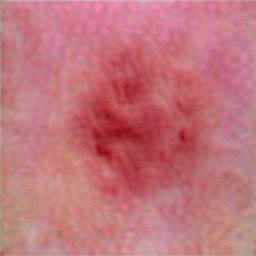}
    \includegraphics[width=0.078\textwidth, height=0.078\textwidth]{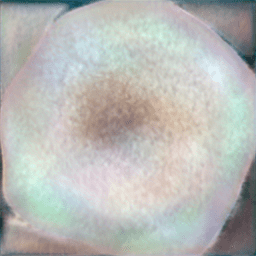}
     \includegraphics[width=0.078\textwidth, height=0.078\textwidth]{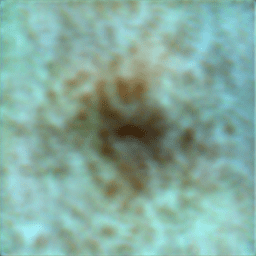}
    \includegraphics[width=0.078\textwidth, height=0.078\textwidth]{figs/isicDefense/Rotation32/malignant/fake_0.png}
}

\subfloat[\normalsize (d) Robust Aggregation]{
    \includegraphics[width=0.078\textwidth, height=0.078\textwidth]{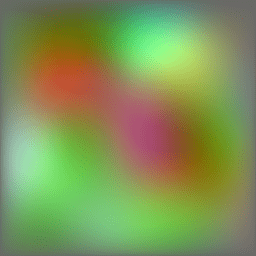}
    \includegraphics[width=0.078\textwidth, height=0.078\textwidth]{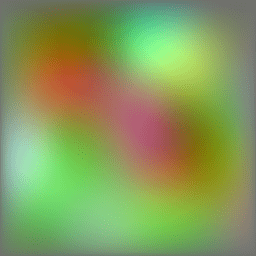}
     \includegraphics[width=0.078\textwidth, height=0.078\textwidth]{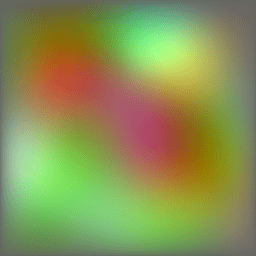}
    \includegraphics[width=0.078\textwidth, height=0.078\textwidth]{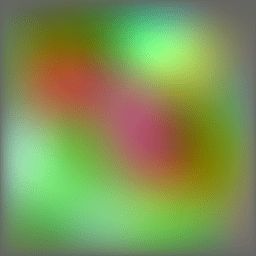}
    
    \includegraphics[width=0.078\textwidth, height=0.078\textwidth]{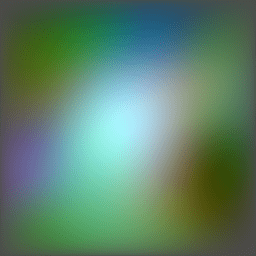}
    \includegraphics[width=0.078\textwidth, height=0.078\textwidth]{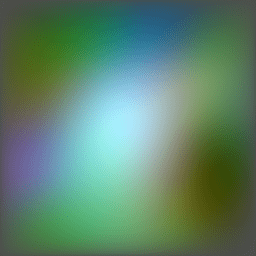}
     \includegraphics[width=0.078\textwidth, height=0.078\textwidth]{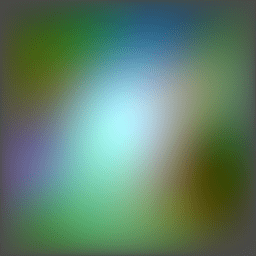}
    \includegraphics[width=0.078\textwidth, height=0.078\textwidth]{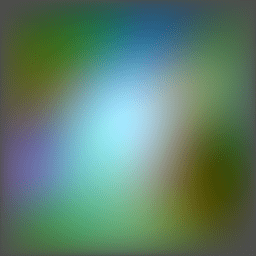}
    
    \includegraphics[width=0.078\textwidth, height=0.078\textwidth]{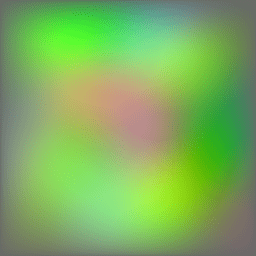}
    \includegraphics[width=0.078\textwidth, height=0.078\textwidth]{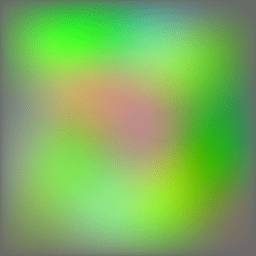}
     \includegraphics[width=0.078\textwidth, height=0.078\textwidth]{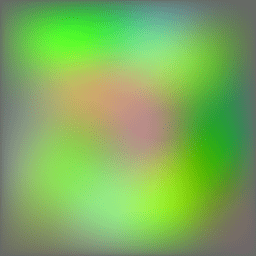}
    \includegraphics[width=0.078\textwidth, height=0.078\textwidth]{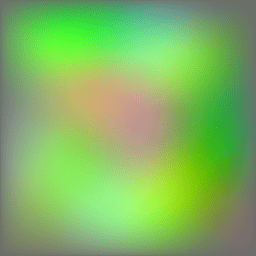}
}
      
      
    \caption{Visualization on different defense strategies for ISIC}
    \label{fig:isicDefense}
\end{figure*}

\begin{table*}[ht]
\centering
\caption{Quantitative Metrics of Defense for ISIC. $\downarrow$ indicates the smaller the better.}

\resizebox{0.99\textwidth}{!}{
\begin{tabular}{@{}cccccccccccccccc@{}}
\toprule
\multirow{3}{*}{Settings} &
\multicolumn{3}{c}{\ours{}~(ours)}&
\multicolumn{3}{c}{Model Reconstruction}&
\multicolumn{3}{c}{Aug: Horizontal Flip}&
\multicolumn{3}{c}{Aug: Random Rotation}&
\multicolumn{3}{c}{\revision{Robust Aggregation}}\\
\cmidrule(l){2-4} 
\cmidrule(l){5-7} 
\cmidrule(l){8-10}
\cmidrule(l){11-13}
\cmidrule(l){14-16}
& 16x16 & 32x32 & 64x64 & 16x16 & 32x32 & 64x64 & 16x16 & 32x32 & 64x64 & 16x16 & 32x32 & 64x64 & \revision{16x16} & \revision{32x32} & \revision{64x64}\\ 
\midrule
LPIPS (B) $\downarrow$ & 0.7337 & 0.6496 & 0.6754 & 0.9420 & 0.8791 & 0.7961 & 0.8504 & 0.9605 & 0.6883 & 0.7566 & 0.8347 & 0.7468 & \revision{0.9696} & \revision{0.9875} & \revision{1.0412}\\
LPIPS (M) $\downarrow$ & 0.8602 & 0.7446 & 0.7502 & 0.8702 & 0.5724 & 0.9553 & 1.028 & 0.9454 & 0.8869 & 0.8198 & 0.8846 & 0.8065 & 0.9717 & 0.9910 & 1.0419\\
\bottomrule
\end{tabular}
}
\label{tab:isicDefense}
\end{table*}
\begin{figure*}[ht]
    \centering
    \captionsetup[subfloat]{labelformat=empty}
    \subfloat[\normalsize \textbf{(a) \ours{} (ours)}]{
    \includegraphics[width=0.99\textwidth]{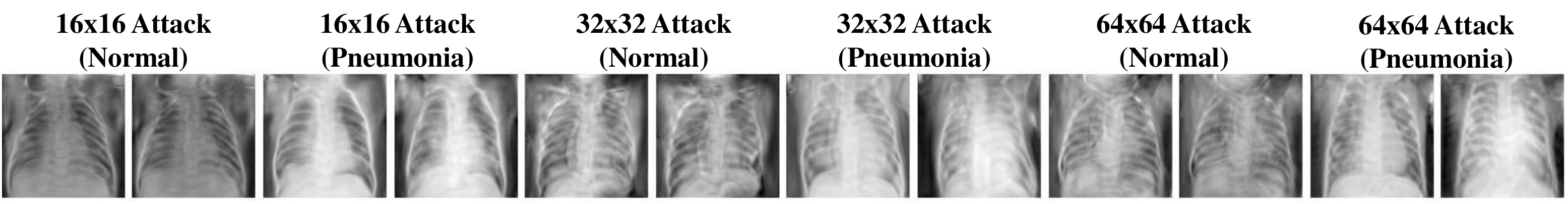}
    }
    
    \subfloat[\normalsize (b) Model Reconstruction]{
    \includegraphics[width=0.078\textwidth, height=0.078\textwidth]{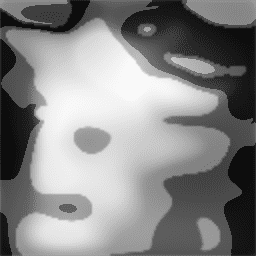}
    \includegraphics[width=0.078\textwidth, height=0.078\textwidth]{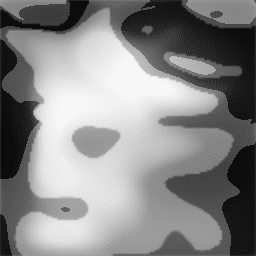}
     \includegraphics[width=0.078\textwidth, height=0.078\textwidth]{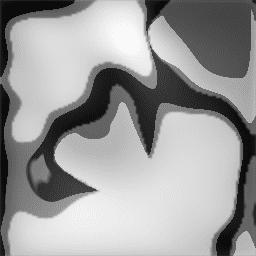}
    \includegraphics[width=0.078\textwidth, height=0.078\textwidth]{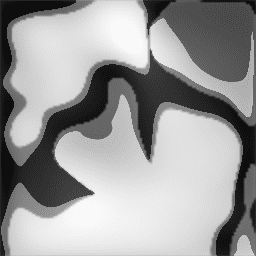}
    
    \includegraphics[width=0.078\textwidth, height=0.078\textwidth]{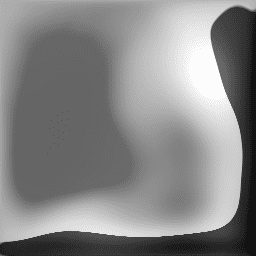}
    \includegraphics[width=0.078\textwidth, height=0.078\textwidth]{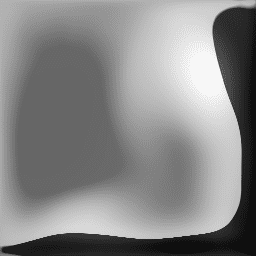}
     \includegraphics[width=0.078\textwidth, height=0.078\textwidth]{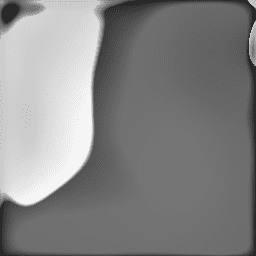}
    \includegraphics[width=0.078\textwidth, height=0.078\textwidth]{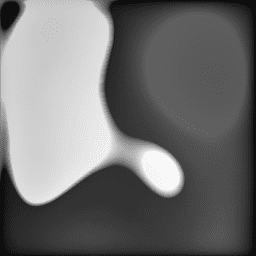}
    
    \includegraphics[width=0.078\textwidth, height=0.078\textwidth]{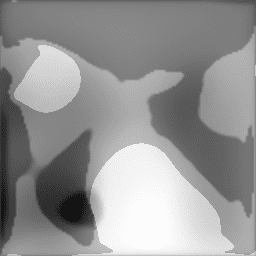}
    \includegraphics[width=0.078\textwidth, height=0.078\textwidth]{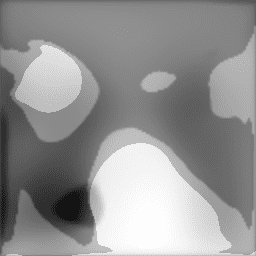}
     \includegraphics[width=0.078\textwidth, height=0.078\textwidth]{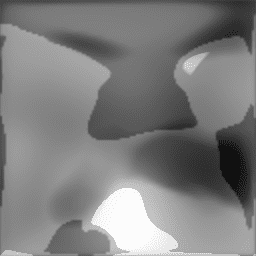}
    \includegraphics[width=0.078\textwidth, height=0.078\textwidth]{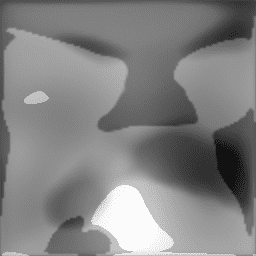}
}

\subfloat[\normalsize \revision{(c) Data Augmentation with Horizontal Flipping abd Random Rotation}]{
    \includegraphics[width=0.078\textwidth, height=0.078\textwidth]{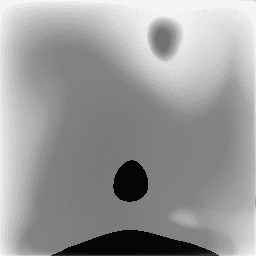}
    \includegraphics[width=0.078\textwidth, height=0.078\textwidth]{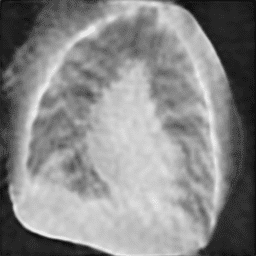}
     \includegraphics[width=0.078\textwidth, height=0.078\textwidth]{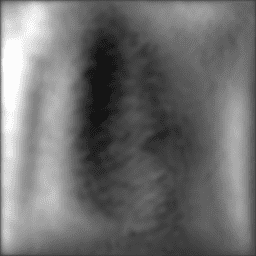}
    \includegraphics[width=0.078\textwidth, height=0.078\textwidth]{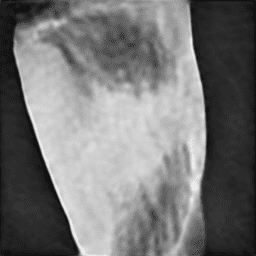}
    
    \includegraphics[width=0.078\textwidth, height=0.078\textwidth]{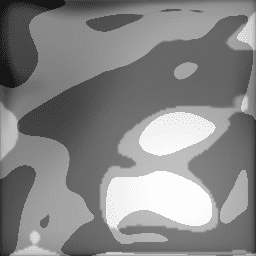}
    \includegraphics[width=0.078\textwidth, height=0.078\textwidth]{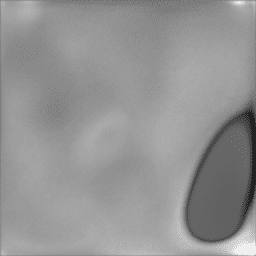}
     \includegraphics[width=0.078\textwidth, height=0.078\textwidth]{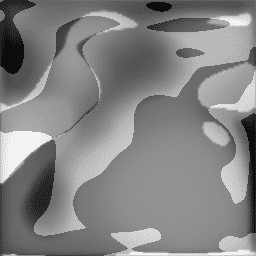}
    \includegraphics[width=0.078\textwidth, height=0.078\textwidth]{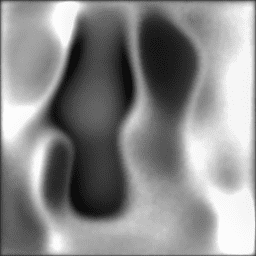}
    
    \includegraphics[width=0.078\textwidth, height=0.078\textwidth]{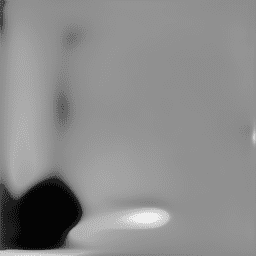}
    \includegraphics[width=0.078\textwidth, height=0.078\textwidth]{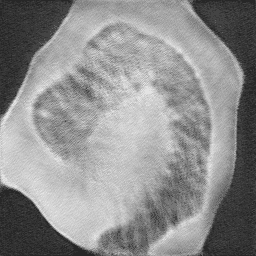}
     \includegraphics[width=0.078\textwidth, height=0.078\textwidth]{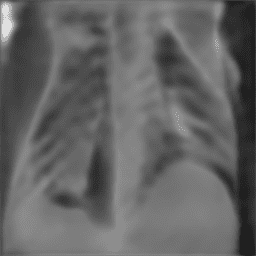}
    \includegraphics[width=0.078\textwidth, height=0.078\textwidth]{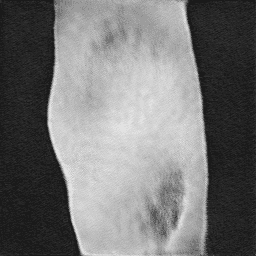}
}

\subfloat[\normalsize \revision{(d) Robust Aggregation}]{
    \includegraphics[width=0.078\textwidth, height=0.078\textwidth]{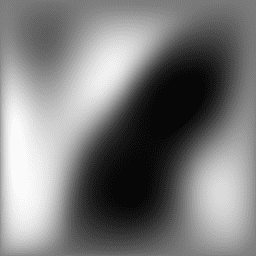}
    \includegraphics[width=0.078\textwidth, height=0.078\textwidth]{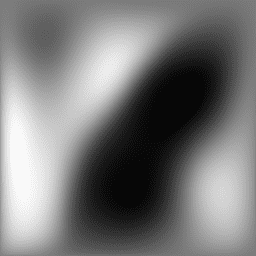}
     \includegraphics[width=0.078\textwidth, height=0.078\textwidth]{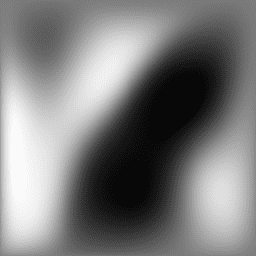}
    \includegraphics[width=0.078\textwidth, height=0.078\textwidth]{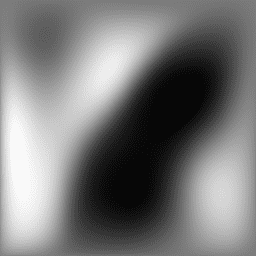}
    
    \includegraphics[width=0.078\textwidth, height=0.078\textwidth]{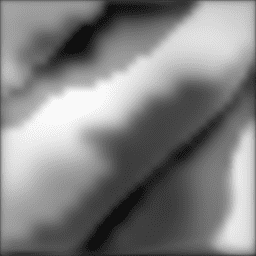}
    \includegraphics[width=0.078\textwidth, height=0.078\textwidth]{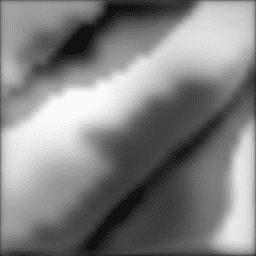}
     \includegraphics[width=0.078\textwidth, height=0.078\textwidth]{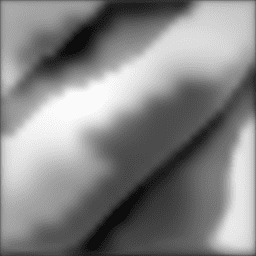}
    \includegraphics[width=0.078\textwidth, height=0.078\textwidth]{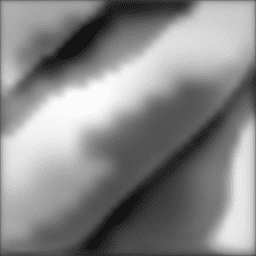}
    
    \includegraphics[width=0.078\textwidth, height=0.078\textwidth]{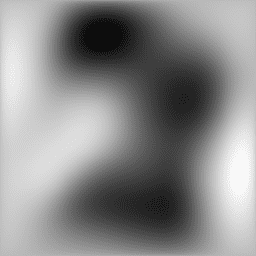}
    \includegraphics[width=0.078\textwidth, height=0.078\textwidth]{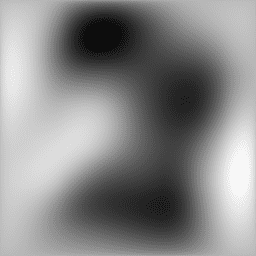}
     \includegraphics[width=0.078\textwidth, height=0.078\textwidth]{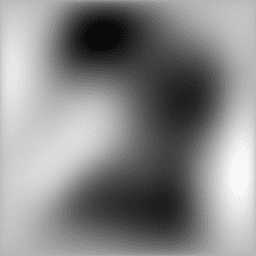}
    \includegraphics[width=0.078\textwidth, height=0.078\textwidth]{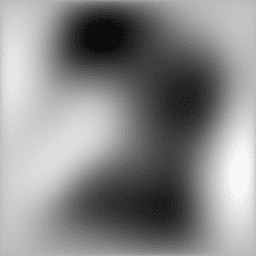}
}
      
      
    \caption{Visualization on different denfense strategies for ChestX}
    \label{fig:chestDefense}
\end{figure*}

\begin{table*}[ht]
\centering
\caption{Quantitative Metrics of Defense for ChestX. $\downarrow$ indicates the smaller the better. }
\resizebox{0.99\textwidth}{!}{
\begin{tabular}{@{}cccccccccccccccc@{}}
\toprule
\multirow{3}{*}{Settings} &
\multicolumn{3}{c}{\ours{}~(ours)}&
\multicolumn{3}{c}{Model Reconstruction}&
\multicolumn{3}{c}{Aug: Horizontal Flip}&
\multicolumn{3}{c}{Aug: Random Rotation}&
\multicolumn{3}{c}{\revision{Robust Aggregation}}\\
\cmidrule(l){2-4} 
\cmidrule(l){5-7} 
\cmidrule(l){8-10}
\cmidrule(l){11-13}
\cmidrule(l){14-16}
& 16x16 & 32x32 & 64x64 & 16x16 & 32x32 & 64x64 & 16x16 & 32x32 & 64x64 & 16x16 & 32x32 & 64x64 & \revision{16x16} & \revision{32x32} & \revision{64x64}\\ 
\midrule
LPIPS (N) $\downarrow$ & 0.4281 & 0.3874 & 0.3892 & 0.6057 & 0.7071 & 0.6154 & 0.7648 & 0.6316 & 0.7767 & 0.5327 & 0.6926 & 0.6311 & \revision{0.7900} & \revision{0.7911} & \revision{0.7897} \\
LPIPS (P) $\downarrow$ & 0.4340 & 0.4187 & 0.3940 & 0.6181 & 0.6813 & 0.6240 & 0.6684 & 0.6346 & 0.6231 & 0.5589 & 0.7137 & 0.6212 & \revision{0.7819} & \revision{0.7875} & \revision{0.7706}\\
\bottomrule
\end{tabular}}
\label{tab:chestDefense}
\end{table*}

\ours{} is applied to the global aggregation step on the server side. To ensure robust detection, recall our outlier detection method described in Algorithm~\ref{alg:cap} requires a warmup process to allow enough time for the malicious clients to overfit the backdoor and behave differently from those benign ones. In our experiments, we set the warmup epoch $m=10$. After $m>10$, generators' losses are required to share with the server to perform malicious detection. For the ISIC dataset, a decay constant $d = 0.9$ is used to penalize weights for the clients detected as an anomaly in every epoch using Isolation Forests~\citep{liu2008isolation}. For ChestX dataset, the decay constant $d = 0.8$ is 
applied in our experiments. We accumulatively count the times of being detected as malicious for each client $c_i(t)$ upon global iteration $t$, at which the calibrated client weights are decayed by timing $d^{c_i(t)}$. Note in the global aggregation, we normalize $w_i$ so that clients' aggregation weights are sum to 1.

\subsubsection{Defense Results and Analysis}

For the ISIC dataset, Fig~\ref{fig:isicDefense} presents the visualization of the defenses experiments and Table~\ref{tab:isicDefense} provides the corresponding quantitative assessment. For the ChestX, Fig~\ref{fig:chestDefense} visualizes the effect of all defense strategies and Table~\ref{tab:chestDefense} presents the corresponding quantitative results.

\noindent\textbf{Visualization Results} As visualized in both row (a) of Fig~\ref{fig:isicDefense} and Fig~\ref{fig:chestDefense}, \ours{} effectively blocks the spread of backdoor attacks from malicious clients and generates images with qualities that are comparable with the vanilla GAN for both ISIC and ChestX in all poisoning trigger sizes.  We also visualize the synthetic images generated with FedGAN with the baseline defense methods described in Section~\ref{exp:moredf}, with the results of ISIC in Fig~\ref{fig:isicDefense} and the results of ChestX in Fig~\ref{fig:chestDefense}. \revision{Specifically, row (b) presents the model reconstruction approach, row (c) presents the data augmentation with horizontal flip and random rotation, and row (d) visualizes generated image robust aggregation with different poison sizes. As can be seen in Fig~\ref{fig:isicDefense}, all the baseline defense strategies fail to block the backdoor attack in FedGAN for ISIC, where the generated images still mismatch the original image.} In ChestX Fig.~\ref{fig:chestDefense}, random horizontal flipping for poison size of 64x64 and random rotation for poison size of 16x16 seems to take effect, which reconstructs the outline of chests. However, this outline is still distorted due to the nature of data argumentation, which may miss important information for diagnostic purposes.

\noindent\textbf{Quantifying Image Fidelity} The corresponding LPIPS score is shown in Table~\ref{tab:isicDefense} for ISIC and Table~\ref{tab:chestDefense} for ChestX. Our defense strategy results in the lowest LPIPS score for almost all entries, in which \ours{} has an average of 0.73 LPIPS score, smaller than the model reconstruction, augmentation, and robust aggregation, where their LPIPS are 0.84, 0.85 and 1.00 individually for ISIC. Similarly for ChestX, \ours{} reaches a score of 0.41, while model reconstruction has an average of 0.64, data augmentation has 0.65 and robust aggregation has the highest average of 0.7851. These are in accordance with what we observed in the visualization plot that \ours{} generates the most similar image as the vanilla FedGAN. Also, the quantitative data for rows (a) matches the observation that LPIPS score recovers to a similar level as the non-attacked group compared to the attacked group in Fig.~\ref{fig:isicAttack} and Fig~\ref{fig:chestAttack}.

\subsection{Synthetic Image Utility Assessment in Classification Tasks}
\label{exp:classification}
As described in Method Section~\ref{rw:fl}, medical data within one medical institution are usually limited, biased, and private, which creates difficulties in its utility, for example, training regular DL models. We simulate the situation of training a classifier only with data in one intuition. The synthetic medical images generated by FedGAN (and their labels) can serve as data augmentation to improve classifier performance. In this part, we trained two classifiers to show the efficacy of the proposed FedGAN pipeline.

\subsubsection{Classifier Architecture and Hyper-Parameter Tuning:} DenseNet121 is applied as the backbone model for training the classifier for both datasets~\citep{huang2017densely}.
All inputs are resized into $224 \times 224$ to fit the default model architecture of DenseNet121. We specify the training details for ISIC and ChestX below.

\noindent\textbf{ISIC: }
The SGD optimizer with a learning rate of $1 \times 10^{-3}$ and weight decay of $1 \times 10^{-4}$ is used for classifying the ISIC images. Also, the learning rate is decreased by $0.9$ after each epoch. The densenet is trained 20 epochs.

\noindent\textbf{ChestX: }
The Adam optimizer with a learning rate of $1 \times 10^{-5}$ is used for training the classifier for the ChestX dataset for 20 epochs. Its learning rate is decayed after each epoch by $0.9$. Also, all input images are augmented by randomly applying horizontal flip and vertical flip.

\subsubsection{Data Augmentation with Synthetic Images in Diagnostic Settings} In order to simulate the real-life challenge of training diagnostic DL models, we set up multiple classifiers and fed each with a different number of real samples. The first experiment simulates training a diagnostic classification model only on limited data (e.g., data within one healthcare institution) while the other mimics FedGAN-facilitated training. In both ISIC and ChestX datasets, the first experiment trains the classifier with real images ranging from 1 to 500 per class sampled from our datasets. Then, we augment the real samples above with 1000 FedGAN-generated synthetic images per class.

\subsubsection{Metrics}
We train the classifiers and evaluate their performance using a pre-processed balanced testing set that contains equal number of samples for each class. In the ISIC dataset, there are 300 Benign samples and 300 Malignant samples. The testing set of ChestX includes 234 Normal images and 234 Pneumonia images. In order to gain robust results, each setting has been executed independently five times with different random seeds. We report the average test accuracy as well as their standard deviation (std).

\subsubsection{Results and Analysis}
\begin{figure*}[tbp]
    \centering
    \subfloat[ISIC]{\label{fig:isicCurve}\includegraphics[width=0.5\textwidth]{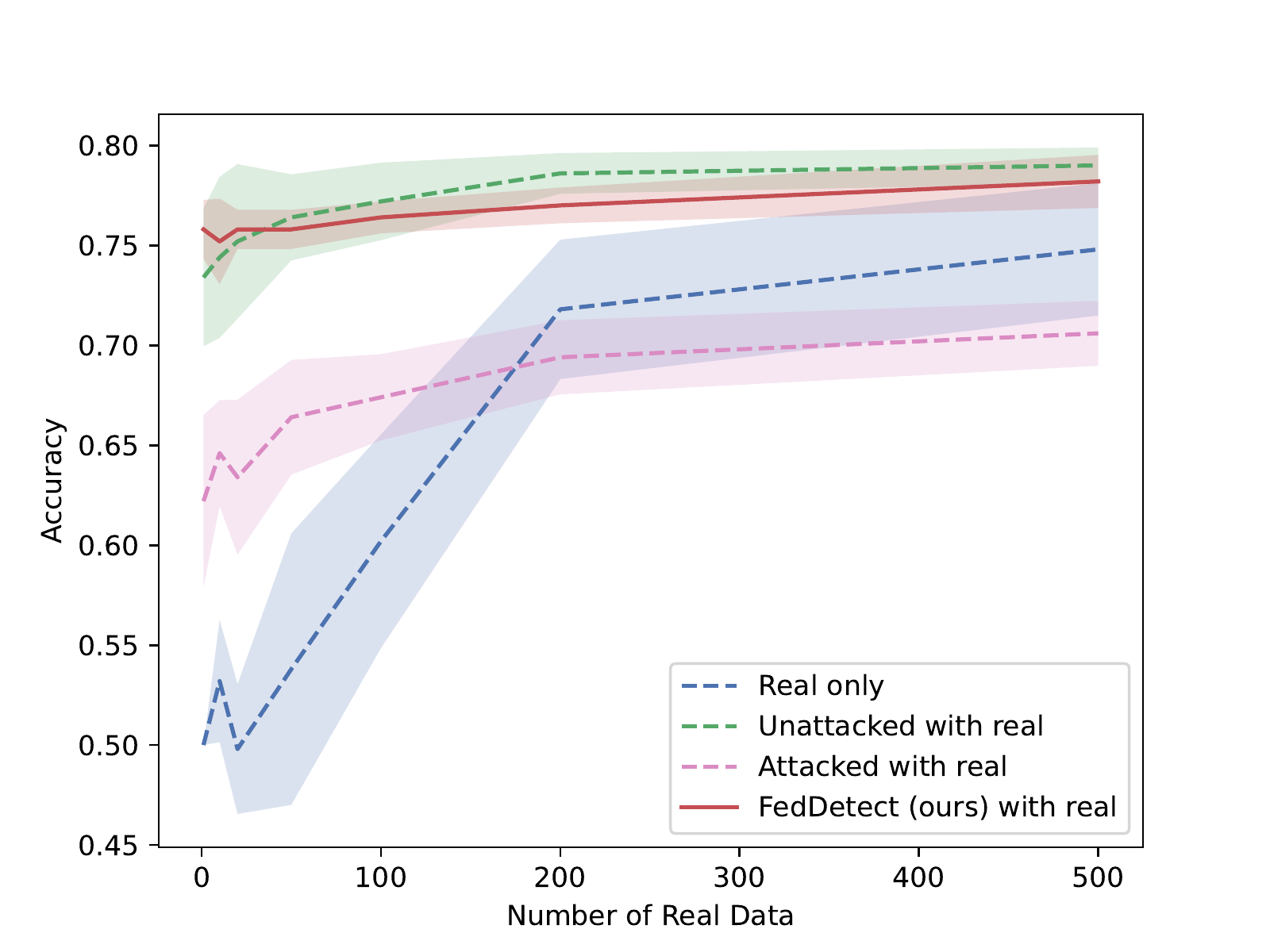}}
    \subfloat[ChestX]{\label{fig:chestXCurve}\includegraphics[width=0.5\textwidth]{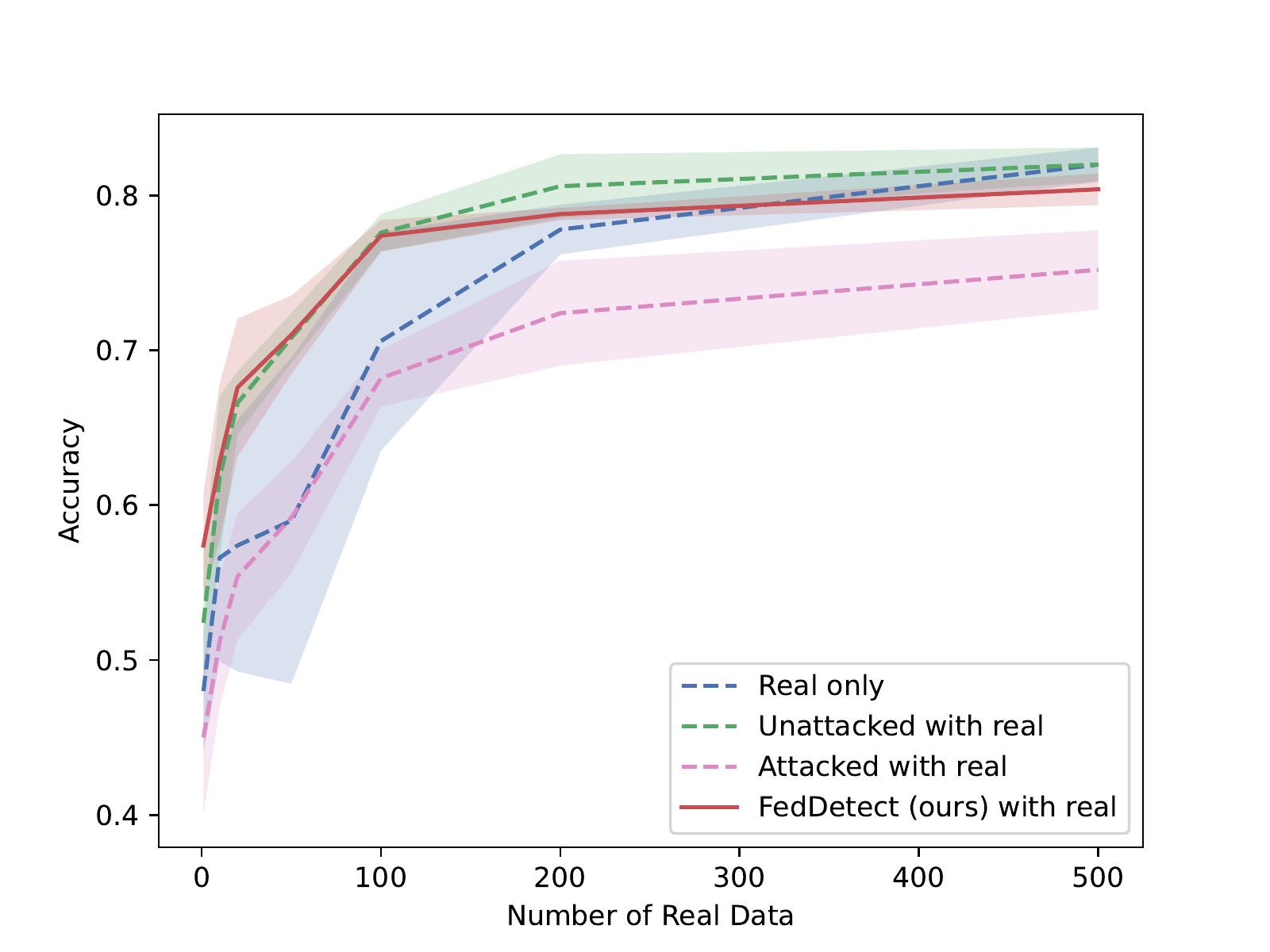}}\\
    \caption{Classification accuracy on synthetic data augmentation with different number of real training data per class.}
    \label{fig:curves}
\end{figure*}


Fig~\ref{fig:isicCurve} and Fig~\ref{fig:chestXCurve} present the test accuracy for both datasets. The blue curve, which is labeled as ``Real only", represents the average test accuracy with only real data. The green line, marked as ``Unattacked with real", represents the average test accuracy of FedGAN-assisted training. The ``Attacked with real" curve represents the FedGAN with backdoor attack with a patch size of 32x32. The red solid curve represents the backdoor attacked FedGAN with \ours{} defense strategy. 
The corresponding shadows stand for the std of the test accuracy over five different trials. The x-axis represents the number of real images added to the training set in one classification trial. For training with all real images, the size of the dataset equals the number on the x-axis. For all other training, the size of the training set equals the number of real images on the x-axis plus the number of generated images from FedGAN, which is 1000 images per class for both datasets.

\noindent\textbf{Utility of High Quality Synthetic Data} As shown in both figures, the augmented dataset using vanilla FedGAN without attack and backdoor attacked FedGAN with \ours{} gains higher test accuracy than purely the real dataset. This is particularly evident for the ISIC dataset, where average test accuracy with the extremely limited training sample (\ie one sample per class) can achieve 0.73 compared to the random guess results of 0.50 from using the real data only. When we have 500 real data per class, adding the synthetic data generated with \ours{} gains an accuracy of 0.80 compared to the accuracy of 0.73 while using real data only.  As for the ChestX dataset, the similar pattern preserves when the real dataset is small (\eg, $< 200$ samples per class). In addition, for both datasets, the average test accuracy using the synthetic data generated with \ours{} under backdoor attack recovers approximately the same as the vanilla FedGAN without attack, which is represented as the green curve in both graphs. The result indicates that synthetic data with high utility can aid diagnostic retains good performance, given it learns the distribution pattern over multiple decentralized datasets, which finally generates a diverse training dataset. This diversity not only benefits training the deep learning classification, but also provides healthcare practitioners more resources to employ in various ways in reality.

\noindent\textbf{Backdoor Attack Affect Data Utility} The effect of augmenting the synthetic data from FedGAN with backdoor attacks is revealed by the orange curves. Seeing from the trend, it lies below all curves for the ChestX dataset, indicating its worst performance. For the ISIC dataset, the attacked images also reach the lowest test accuracy when the real images are greater than 200, which indicates augmenting with low-quality synthetic data from the poisoned generative model even can hurt classification performance. 

The classification results corresponds to visualization in the previous Sections~\ref{exp:attack} and~\ref{exp:defence}. The attack corrupts the overall performance of FedGAN and \ours{} effectively suppresses the adversarial behavior.

\section{Conclusion}
Motivated by the idea of backdoor attacks in classification models, this work investigates the pitfalls of backdoor attacks in training conditional FedGAN models. We conduct extensive experiments to investigate the backdoor attack on two public datasets and evaluate among different types and sizes of triggers. Based on our key observations on malicious clients' loss patterns, we propose \ours{} as an effective defense strategy against backdoor attacks in FedGAN. We comprehensively conduct quantitative and qualitative assessments on the fidelity and utility of the synthetic images under different training conditions. We demonstrate the \ours{} significantly outperforms the alternative baselines and preserve comparable data utility as attack-free vanilla FedGAN.

As the first step towards understanding backdoor attacks in FedGAN for medical image synthesis, our work brings insight into building a robust and trustworthy model to advance medical research with synthetic data. Furthermore, we hope to highlight that \ours{} involves only lightweight improvement on the server aggregation step. This makes \ours{} flexible to integrate into different GAN-based federated generative models. Our future work includes scaling up the FL system with more clients, generalizing \ours{} to other deep generative models, \eg, Diffusion models~\citep{song2020score}, and considering other variants of backdoor attacks, \eg, frequency-injection based attack~\cite{feng2022fiba}.

\section*{Acknowledgments}
This work is supported in part by the Natural Sciences and Engineering Research Council of Canada (NSERC), Public Safety Canada (NS-5001-22170), and NVIDIA Hardware Award. We thank Chun-Yin Huang and Nan Wang for their kind instruction to Ruinan Jin and assistance with implementation.





\bibliographystyle{model2-names.bst}\biboptions{authoryear}
\bibliography{refs}

\end{document}